%% file: main.tex
\documentclass[10pt,twocolumn]{article}

\usepackage[margin=1in]{geometry}
\usepackage{graphicx}
\usepackage{amsmath,amssymb,amsfonts}
\usepackage{textcomp}
\usepackage{algorithm}
\usepackage{algpseudocode}
\usepackage{booktabs}
\usepackage{multirow}
\usepackage{makecell}
\usepackage{subcaption}
\usepackage[table]{xcolor}
\usepackage[framemethod=tikz]{mdframed}
\usepackage{soul}
\usepackage[numbers]{natbib}
\usepackage[hidelinks]{hyperref}

\newif\ifrevision
\revisionfalse 

\ifrevision
    \definecolor{lightyellow}{RGB}{255,245,185}
    \definecolor{revgreen}{RGB}{0,120,0}
    \sethlcolor{lightyellow}
    
    \DeclareRobustCommand{\rev}[1]{\hl{#1}} 
    
    \newcommand{\revterm}[1]{%
      \begingroup
      \setlength{\fboxsep}{0.4pt}%
      \colorbox{lightyellow}{\strut #1}%
      \endgroup
    } 
    
    \newcommand{\revshort}[1]{%
      \begingroup
      \setlength{\fboxsep}{0.5pt}%
      \colorbox{lightyellow}{\strut #1}%
      \endgroup
    }

    \newcommand{\revframe}[1]{%
        \setlength{\fboxrule}{2pt}%
        \fcolorbox{yellow}{white}{#1}%
    }
    
    \newcolumntype{R}{>{\columncolor{yellow!15}}c}
\else
    \newcommand{\rev}[1]{#1}
    \newcommand{\revterm}[1]{#1}

    \newcommand{\revshort}[1]{#1}
    \newcommand{\revframe}[1]{#1}

    \newcolumntype{R}{c}
\fi

\title{Limited-Angle Tomography Reconstruction via Projector Guided 3D Diffusion}
\date{}

\author{
    Zhantao Deng,
    Mériem Er-Rafik,
    Anna Sushko,
    Cécile Hébert, 
    Pascal Fua\thanks{
    Zhantao Deng, Pascal Fua are with the School of Computer and Communication Sciences, École Polytechnique Fédérale de Lausanne, 1015 Lausanne Switzerland. Mériem Er-Rafik, Cécile Hébert are with the School of Basic Sciences, École Polytechnique Fédérale de Lausanne, 1015 Lausanne Switzerland. Anna Sushko is with the School of Computer and Communication Sciences and also with the School of Basic Sciences, École Polytechnique Fédérale de Lausanne, 1015 Lausanne Switzerland.
    \texttt{E-mail: firstname.lastname@epfl.ch}.
    Corresponding author: Zhantao Deng. \\
    We are grateful to Prof. Graham Knott (The School of Life Sciences, École Polytechnique Fédérale de Lausanne) for providing access to the FIB-SEM data and thin sections used in this work, and for granting permission to use these data and samples for our analyses.\\
    The code and data used in this study will be made publicly available upon publication of this article.
    }
}

\begin{document}
\maketitle

\input{tex/defs.tex}

\input{tex/abstract}

\input{tex/intro}
\input{tex/related}

\input{tex/method}
\input{tex/material}
\input{tex/results}
\input{tex/conc}

\bibliography{bib/string,bib/refs}
\bibliographystyle{plainnat}

\end{document}

%% file: tex/defs.tex

%
%
\newif\ifdraft
\draftfalse
\drafttrue

\ifdraft
 \newcommand{\pf}[1]{{\color{red}{#1}}}
 \newcommand{\PF}[1]{{\color{red}{{\bf PF}: \bf #1}}}
 \newcommand{\zd}[1]{{\color{blue}{#1}}}
 \newcommand{\ZD}[1]{{\color{blue}{{\bf ZD}: #1}}}
 \newcommand{\ch}[1]{{\color{cyan}{#1}}}
 \newcommand{\CH}[1]{{\color{cyan}{\bf CH: #1}}}
 \newcommand{\as}[1]{{\color{cyan}{#1}}}
 \newcommand{\AS}[1]{{\color{cyan}{AS: #1}}}
 \newcommand{\mer}[1]{{\color{green}{#1}}}
 \newcommand{\MER}[1]{{\color{green}{\bf MER: #1}}}
 \newcommand{\TODO}[1]{{\textcolor{red} {TODO: #1}}}
 \newcommand{\ONGOING}[1]{{\textcolor{orange} {ONGOING: #1}}}
\else
 \newcommand{\pf}[1]{#1}
 \newcommand{\zd}[1]{#1}
 \newcommand{\ch}[1]{#1}
 \newcommand{\as}[1]{#1}
 \newcommand{\PF}[1]{}
 \newcommand{\ZD}[1]{}
 \newcommand{\CH}[1]{}
 \newcommand{\AS}[1]{}
 \newcommand{\TODO}[1]{}
 \newcommand{\ONGOING}[1]{}
\fi

\newcommand{\comment}[1]{}
\newcommand{\parag}[1]{\vspace{-0.9mm}\paragraph{#1}}
\newcommand{\sparag}[1]{\vspace{-0.9mm}\subparagraph{#1}}
\newcommand{\subsec}[1]{\vspace{-0.4mm}\subsection{#1}}

%
%
\newcommand{\ours}[0]{\textit{TEMDiff}}
\newcommand{\mbir}[0]{\textit{DiffusionMBIR}}
\newcommand{\dolc}[0]{\textit{DOLCE}}
\newcommand{\scd}[0]{\textit{SCD}}
\newcommand{\ddip}[0]{\textit{D3IP}}
\newcommand{\pfitre}[0]{\textit{PFITRE}}
\newcommand{\blaze}[0]{\textit{BLAZE3DM}}

%
%
\newcommand{\figy}[2]{\includegraphics[height=#1,keepaspectratio]{#2}}
\newcommand{\figx}[2]{\includegraphics[width=#1,keepaspectratio]{#2}}

%
%

\newcommand{\fdyn}[0]{{f_{\rm dyn}}}
\newcommand{\gdyn}[0]{{g_{\rm dyn}}}
\newcommand{\gcnn}[0]{{\tilde{g}_{\rm dyn}}}

\newcommand{\mL}[0]{\mathcal{L}}
\newcommand{\mM}[0]{\mathcal{M}}
\newcommand{\mN}[0]{\mathcal{N}}
\newcommand{\mR}[0]{\mathcal{R}}
\newcommand{\mS}[0]{\mathcal{S}}
\newcommand{\mU}[0]{\mathcal{U}}
\newcommand{\mZ}[0]{\mathcal{Z}}

\newcommand{\bA}[0]{\mathbf{A}}
\newcommand{\bc}[0]{\mathbf{c}}
\newcommand{\bl}[0]{\mathbf{\Lambda}}
\newcommand{\bt}[0]{\mathbf{\Theta}}
\newcommand{\bC}[0]{\mathbf{C}}
\newcommand{\bI}[0]{\mathbf{I}}
\newcommand{\bP}[0]{\mathbf{P}}
\newcommand{\bQ}[0]{\mathbf{Q}}
\newcommand{\bR}[0]{\mathbf{R}}
\newcommand{\br}[0]{\mathbf{r}}
\newcommand{\bs}[0]{\mathbf{s}}
\newcommand{\bu}[0]{\mathbf{u}}
\newcommand{\bU}[0]{\mathbf{U}}
\newcommand{\bn}[0]{\mathbf{n}}
\newcommand{\bN}[0]{\mathbf{N}}
\newcommand{\bm}[0]{\mathbf{m}}
\newcommand{\bp}[0]{\mathbf{p}}
\newcommand{\bv}[0]{\mathbf{v}}
\newcommand{\bX}[0]{\mathbf{X}}
\newcommand{\bx}[0]{\mathbf{x}}
\newcommand{\by}[0]{\mathbf{y}}
\newcommand{\bY}[0]{\mathbf{Y}}
\newcommand{\bw}[0]{\mathbf{w}}
\newcommand{\bW}[0]{\mathbf{W}}
\newcommand{\bz}[0]{\mathbf{z}}
\newcommand{\bZ}[0]{\mathbf{Z}}

%% file: tex/abstract.tex

\begin{abstract}
Limited-angle electron tomography aims to reconstruct 3D shapes from 2D projections of Transmission Electron Microscopy (TEM) within a restricted range and number of tilting angles, but it suffers from the missing-edge problem that causes severe reconstruction artefacts. Deep learning approaches have shown promising results in alleviating these artefacts, yet they typically require large high-quality training datasets with known 3D ground truth, which are difficult to obtain in electron microscopy. To address these challenges, we propose \ours{}, a novel 3D diffusion-based iterative reconstruction framework. Our method is trained on readily available volumetric FIB-SEM data using a simulator that maps them to TEM tilt series, enabling the model to learn realistic structural priors without requiring clean TEM ground truth.
By operating directly on 3D volumes, \ours{} implicitly enforces consistency across slices without the need for additional regularisation. In simulated electron tomography datasets with limited angular coverage, \ours{} outperforms the most advanced methods in reconstruction quality. We further demonstrate that a trained model \ours{} generalises well to real-world TEM tilts obtained under different conditions and can recover accurate structures from tilt ranges as narrow as 8 degrees, with 2-degree increments, without any retraining or fine-tuning.
\end{abstract}

%% file: tex/intro.tex

\section{Introduction}

\begin{figure}[t]
\centerline{\includegraphics[width=\columnwidth]{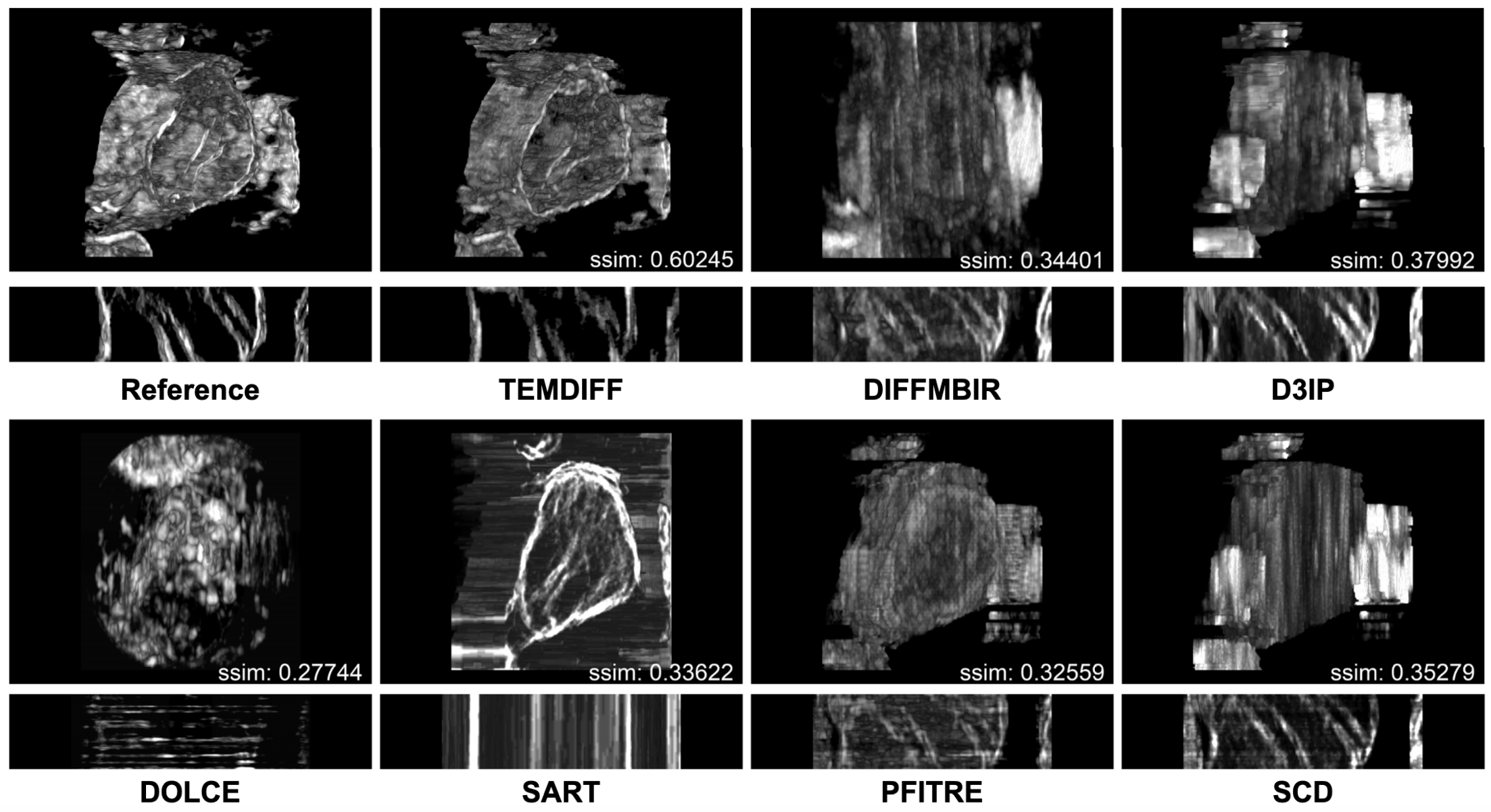}}
\caption{Mitochondria reconstructed using \revshort{\ours{}, \mbir{}, \ddip{}, \dolc{}, SART,} \revshort{\pfitre{}, \scd{}}, with tilts covering $10^\circ$ with $1^\circ$ increments. The top row presents 3D views, while the bottom row displays vertical cross-section along x axis of each corresponding 3D view at the volume center. \ours{} produces more clear, more consistent and better structures than others in both views. The contrast for these other methods has been enhanced for better visual quality.}
\label{fig:teaser}
\end{figure}

Electron tomography (ET) is used to reconstruct a three-dimensional ultrastructure from two-dimensional projections acquired at multiple tilt angles. It has become an indispensable tool in structural biology and materials science, providing nanometer-scale insights into organelles, synapses, and complex assemblies within individual cells in both healthy and diseased states~\cite{Glancy23}. In principle, accurate reconstructions require tilt ranges approaching $180^{\circ}$ with fine angular increments~\cite{Frank06, Barcena09, Ercius15}. In practice, however, mechanical limits of the microscope, specimen geometry, and electron dose constraints often restrict the angular range, necessitating limited-angle computed tomography (LACT) with a tilt range much smaller than $180^{\circ}$. As the tilt range narrows, LACT becomes increasingly challenging. In fact, when it drops to $10^\circ$ or less, even the best current methods\rev{, such as }\revshort{\mbir{}~\cite{Chung23a}, \ddip{}~\cite{Chung24}, \dolc{}~\cite{Liu23k}, SART~\cite{Zheng22a}, \pfitre{}~\cite{Zhao25}, \scd{}~\cite{webber25},}\rev{ struggle to recover meaningful results as demonstrated in}\revshort{ Fig.~\ref{fig:teaser}.}

\rev{ These difficulties arise primarily from incomplete Fourier coverage, commonly referred to as the missing-wedge problem. According to the Fourier slice theorem, each projection contributes information along a plane in the Fourier domain. When the angular range is restricted, large regions of Fourier space remain unobserved, forming a wedge-shaped region of missing frequencies. This results in anisotropic resolution and introduces reconstruction artifacts such as elongation along the beam direction, streaking, and loss of structural fidelity. Their severity  increases rapidly as the angular range decreases, making extremely limited-angle tomography a severely ill-posed inverse problem. This has been extensively studied and analyzed in both the classical tomography}\revshort{~\cite{Kak01, Wang23d}}\rev{and electron microscopy literature}\revshort{~\cite{Ercius15, Midgley03}}.

In spite of these difficulties, the pursuit of tomography with such limited tilt ranges has several advantages. First, acquiring fewer images reduces the total electron dose and shortens the acquisition time, which is critical for preserving radiation-sensitive biological structures~\cite{Frank06, Grant15}. 
Second, a narrow tilt range improves data quality because large tilt angles increase the effective specimen thickness, leading to enhanced multiple scattering and reduced image contrast. In addition, mechanical positioning errors, drift, and defocus gradients become more pronounced at high tilts, resulting in projection distortions and reduced alignment accuracy~\cite{Mastronarde05}.
 As a result, individual projection images are of higher and more uniform quality, improving alignment accuracy and reconstruction stability. In addition, narrow tilt ranges make tomography feasible for thick, fragile, or geometrically constrained specimens that cannot tolerate large angular excursions~\cite{Midgley03, Leis09}. In short, a robust reconstruction method for extremely limited angular coverage would significantly expand the applicability of electron tomography to experimental conditions currently inaccessible with conventional acquisition strategies.

\begin{figure*}[t]
\centerline{\includegraphics[width=1\textwidth]{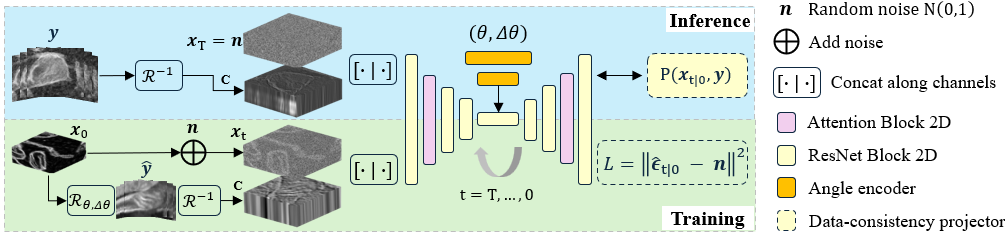}}

\caption{The \ours{} pipeline.
{\bf (Top)} At inference time, given the acquired tilts $\by$, the inverse radon transform $C=\mR^{-1}(\by)$ is computed and concatenated with noise. The result is fed to the pretrained U-Net $\mN$ which includes attention layers and takes the acquisition angular range $\theta$ and increments $\Delta\theta$ as further inputs. The U-Net $\mN$ estimates the amount of noise in the input and iteratively denoise it. 
{\bf (Bottom)} To train $\mN$ without access to ground-truth data, FIB-SEM data is leveraged to synthesize realistic STEM tilts $\Hat{\by}$, given $\theta$ and $\Delta\theta$. Then, the inverse radon transform $C=\mR^{-1}(\Hat{\by})$ is computed and concatenated with the noised ground truth patch. The result is fed to $\mN$ together with $\theta$ and $\Delta\theta$ to predict the noise in the input, which is used to train $\mN$ by minimizing the mean squared error.
}

\label{fig:full_pipeline}
\end{figure*}

To address the need for robust reconstruction from extremely limited tilt ranges down to about 10 degrees, we introduce \ours{}, a projector-guided 3D conditional diffusion framework tailored to LACT~\cite{Ho22a, saharia2022, Liu23k}. Our method focuses on reconstructing biological microstructures imaged by (scanning) transmission electron microscopy ((S)TEM), with a particular emphasis on mitochondria and synapses. Mitochondria and synapses are among the most easily identifiable organelles in electron microscopy images~\cite{Glancy23} and play critical roles in cellular energy conversion, signaling, and biosynthetic processes. However, these organelles exhibit complex morphologies and a wide range of sizes, and differences in specimen preparation introduce further variability. This diversity increases the data requirements for training a neural network model. At the same time, artifact-free ground-truth volumes are nearly impossible to obtain, since even the best tomographic reconstructions are affected by the missing wedge problem.

Given these challenges, our approach differs from prior diffusion-based tomography methods in several key ways. Unlike methods that operate on 2D slices to mitigate data shortage and require explicit enforcement of inter-slice consistency, or latent-space approaches that learn an implicit manifold~\cite{He25, Liu25} and demand large numbers of high-quality volumes, \ours{} is trained directly on 3D volumes. This end-to-end volumetric training allows the model to naturally capture inter-slice correlations and maintain cross-slice consistency without additional constraints. Furthermore, to address the lack of artifact-free training data, we synthesize tilt-series data from FIB-SEM volumes using a physics-motivated simulator, ensuring that our training data are statistically close to real measurements. Finally, to keep the diffusion output consistent with the actual measurements, we incorporate a projection-based correction step at each denoising iteration~\cite{Chung23a, Zheng22a, Chung22a}. This strategy, inspired by previous work, enforces data fidelity in a simple, efficient, and accurate manner, \revshort{as depicted in Fig.~\ref{fig:full_pipeline}}.

As a result, \ours{} produces qualitatively better reconstructions than conventional methods such as the SART algorithm, as well as recent state-of-the-art approaches, especially when the tilt range is on the order of $10^\circ$ or less. This performance far exceeds that of other methods, which typically require tilt ranges of $60^\circ$–$100^\circ$. \rev{A reconstruction obtained with }\revterm{\ours{}}\rev{ using a $10^\circ$ tilt range is also presented in Fig.}\revshort{~\ref{fig:teaser}}\rev{ for comparison.} Importantly, these results are achieved using only the synthetic training data described above, without any fine-tuning on real data. This demonstrates the practicality of our approach in real-world settings, since obtaining artifact-free ground-truth volumes is nearly impossible in electron tomography.

%% file: tex/related.tex

\section{Related Work}
\label{sec:related}
\rev{Tomographic reconstruction aims to recover a three-dimensional volume from a set of projection images acquired at several tilt angles. In practical electron microscopy experiments,
however, full angular coverage is not attainable due to physical constraints such as specimen thickness, stage geometry, and electron dose limitations. Consequently, reconstruction must often be performed using only a severely limited range of angles, which amounts to only partially sampling the Fourier domain. This gives rise to the missing-wedge problem, resulting in reconstruction artifacts and reduced structural fidelity. To mitigate these effects, a broad range of reconstruction strategies have been proposed, including analytical methods, model-based iterative reconstruction techniques, and, more recently, deep learning based approaches.}

\subsection{Conventional Methods}

\subsubsection{Filtered Back Projection (FBP)}

This classic analytical method relies on the Fourier slice theorem and is computationally efficient. However, it requires an almost complete angular range (close to $180^\circ$ or more) for accurate reconstruction. With narrower coverage, FBP reconstructions are especially vulnerable to missing-wedge artifacts, producing streaking and anisotropic elongation~\cite{Kak01,Wang23c}. While weighting and other corrections can mitigate these effects~\cite{zeng12,Willemink19}, no analytic method can fully overcome the severe ill-posedness of LACT.

\subsubsection{Model-Based IR (MBIR)}

By modeling the imaging process using a system of linear equations, LACT can be formulated as an inverse problem and solved using iterative optimization techniques~\cite{Kak01,Geyer15}. Regularizers impose assumptions such as sparsity~\cite{Aharon06}, smoothness~\cite{Zhan17e} or shape priors ~\cite{Xia23a, Rond16} and the solution is often cast as a maximum a posteriori (MAP) estimation problem. MBIR is generally more robust than analytic methods and performs better under limited-angle or low-dose conditions~\cite{Willemink19}. However, when the tilt range is extremely restricted, the problem becomes so under-determined that handcrafted constraints fail to recover the missing information~\cite{Wang23e}.

\subsection{Deep Learning Methods}
Deep learning has been extensively investigated to provide more powerful data-driven priors than those the MBIR hand-crafted regularization terms provide, yielding  improvements under challenging conditions~\cite{Willemink19}. One approach is end-to-end reconstruction, mapping directly from projection data to volume ~\cite{kishore25, Khorashadizadeh25, Vahteristo25} with regularization to enforce consistency. Another is post-processing, where networks denoise and refine intermediate reconstructions ~\cite{Liu20h, Yao24, Wiedemann24}. More commonly, deep learning is integrated within traditional reconstruction frameworks such as FBP or MBIR, to combine the strengths of data-driven priors with those of physics-based modeling. This can be done by either unrolling or regularization. In unrolling-based methods, the iterative reconstruction algorithm is “unrolled” into a fixed sequence of steps, and trainable neural network layers are introduced to solve each one sequentially~\cite{Chen18h, Adler18}. In regularization-based methods, a deep network is used as a trainable denoiser that injects prior knowledge into the intermediate reconstruction steps, while data fidelity is enforced by the physics-based models~\cite{Chung23a,Chung22a, Zhao25}. This has been shown to improve reconstruction quality in limited-angle and low-dose cases, compared to purely analytical methods. Nevertheless, it remains to be further investigated how to keep deep learning methods flexible and robust while the scenario is heavily ill-posed with missing-information and there is limited data for training.

\subsection{Diffusion Based Methods} \label{sec:diffusion}
Diffusion models have recently emerged as powerful generative priors, capable of producing high-quality images by iteratively refining noise into structured outputs~\cite{Song21c, Ho20a}. The sampling process can be guided by physical constraints, making it appealing for limited-angle tomographic reconstruction.

To incorporate measurement constraints into diffusion-based reconstruction, a variety of strategies have been explored. For example, Song et al.~\cite{Song21d} propose guiding an unconditional diffusion model by adding a proximal optimization step with noised measurements at each reverse diffusion iteration. Others methods, like \dolc{} generates each slice conditioned on its initial reconstruction from the available projections. TIFA~\cite{Wang24} accelerates sampling by combining large “jumps” in the reverse diffusion process with intermediate re-sampling; at each step, measured projections correct the output and a diagonal total variation regularizer stabilizes the jumps. PWD~\cite{Liu25b} adds a wavelet-transformation module and a Gaussian prior for guidance in the diffusion model, achieving faster sampling with high-fidelity reconstructions. Similarly, PSDM~\cite{Han24} couples the diffusion process with a primal-dual hybrid gradient algorithm, running inner iterations at every reverse step to enforce data fidelity.

Several advanced strategies further improve fidelity and consistency in 2D. RN-SDEs~\cite{Guo24c} integrates range-null space decomposition to mean-revert diffusion as rectification. DPER~\cite{Du24} embeds the diffusion prior into an implicit neural representation that is optimized for measurement consistency; the refined output is then noised and fed back into the next diffusion iteration. CSN~\cite{Zhang25d} recovers missing information in the projection null space via a conditional diffusion model and fuses it with the measured range-space reconstruction, preserving fidelity. MISD~\cite{Wang25} addresses measurement noise by introducing a weighted least-squares data fidelity term and a noise-suppression regularizer into the reverse diffusion process.

Some approaches exploit alternative domains or multi-domain priors. For instance, Guo et al.~\cite{Guo25} train a diffusion model directly on the sinogram domain and perform reconstruction as an inpainting task; they distill this sinogram prior for faster inference and apply post-processing to refine the output. WISM~\cite{Zhang24} alternates between wavelet-domain and image-domain diffusion to improve reconstruction quality. PEDB~\cite{Wang25a} uses a diffusion bridging model, explicitly injecting the measured projections at each reverse step to enforce fidelity throughout the sampling process.

To ensure cross-slice consistency and extend diffusion models to 3D volumes, researchers have integrated diffusion priors into volumetric reconstruction pipelines. \mbir{} incorporates a pre-trained 2D diffusion prior into a model-based 3D iterative reconstruction framework: the data fidelity term enforces consistency with measured projections, while an explicit regularization encourages adjacent slices to agree. Blaze3DM~\cite{He25} and D3T~\cite{Liu25} embed diffusion priors within tri-plane representations of 3D volumes. Blaze3DM trains a diffusion model on a tri-plane feature representation of the volume and guides it at each reverse step with the available measurements. D3T compresses 3D volumes into a tri-plane latent space via a vector-quantized autoencoder, then performs conditional diffusion in both the projection domain and the image domain.

Recognizing the scarcity of high-quality training data in medical imaging, some works focus on test-time adaptation and domain transfer. \scd{} trains a diffusion prior on synthetic data and then uses a LoRA-based~\cite{Hu22} fine-tuning for each slice at test time on real data to bridge the distribution gap. \ddip{} improves upon \scd{} by reusing the adapted parameters across slices and jointly adapting multiple slices, rather than tuning each slice independently. In the context of limited-angle cone-beam computer tomography, LARGEDiffNet~\cite{Wu25} guides a diffusion model using a corresponding planning computer tomography (CT) scan of the patient, which is registered to the target volume via unsupervised translation and registration models, to provide additional prior information.

Even though these methods achieve state-of-the-art results on moderate tilt ranges (e.g., $90^\circ$ or $120^\circ$), they often incur heavy computational costs, slow inference, or require a large amount of high-quality training data. In narrow limited-angle scenarios (e.g. $10^\circ$), even advanced data-free approaches with test-time training suffer severe degradation. With so few projections, the reconstruction problem becomes highly ill-posed: numerous distinct 2D slice configurations can satisfy both the diffusion prior and the sparse measurements, and typical per-slice regularizations (implicit or explicit) cannot enforce a globally consistent solution. As a result, slice-by-slice diffusion processes may produce individually plausible slices that fail to form a coherent 3D volume, or yield a coherent 3D volume that matches the object's shape but with poor slice-wise quality. Distortions or misalignments in the tilt series can further worsen cross-slice inconsistencies. 

Furthermore, latent-space approaches struggle to recover fine details under these conditions, due to the sparse data and the limited precision of the learned manifolds. These challenges motivate our proposed method, \ours{}. By operating directly on the 3D volume, \ours{} conditions each slice not only on its own measurements but also on all other slices and their measurements. In other words, the learned prior is applied in the full volumetric (pixel) space rather than independently per slice. This coordinated 3D approach reduces the ill-posedness of the problem and yields reconstructions with improved cross-slice consistency and overall quality.

%% file: tex/method.tex
\section{Reconstruction Method} \label{sec:method}

We address LACT by coupling a 3D conditional diffusion prior with an uncertainty-weighted data-consistency corrector. The prior is learned from volumetric FIB-SEM data using a physics-inspired simulator that maps volumes to STEM HAADF tilt series, enabling the transfer of structural priors without 3D ground truth. This is a crucial step because artifact-free ground truth is virtually unobtainable from (S)TEM images: All reconstructions suffer from the missing wedge problem and sparse angular sampling, both inherent consequences of the mechanical design of electron microscopes. In this section, we first describe the guided reconstruction framework and then present the contrast model used for dataset preparation.

\subsection{Guided Reconstruction} \label{sec: la_tomo}

Given an unknown density $\bx \in \mathbb{R}^{D \times H \times W}$, the Radon transform $\mR_{\theta, \Delta\theta}(\bx)$ is the tilting series obtained as the output of a tomographic scan within angular range $\theta$ and increment $\Delta \theta$. It is also known as a sinogram $\by$. For brevity, we omit the subscript and write \(\mR(\bx)\), with the dependence on ($\theta,\Delta\theta$) understood from context. The Radon transform is a linear transform of $\bx$ and can be written as
\begin{equation}\label{eq:radon}
   \by = \mR(\bx) = \bA \bx \; ,
\end{equation}
where the matrix $\bA$ models a line integration operation over the set of tilting angles defined by angle range $\theta$ and step $\Delta\theta$. If the tilt angles range from $0$ to $180^{\circ}$ with a sufficiently small $\Delta\theta$, the inverse Radon transform $ \mR^{-1}(\by)$ can be computed and yields a good estimate $\hat \bx$ of the original $\bx$. In LACT, this is not the case because the range is reduced and~\eqref{eq:radon} has many solutions. This can be handled by finding the most likely $\hat \bx$ that satisfies~\eqref{eq:radon}, in other words solving a Maximum A Posteriori (MAP) Estimation problem. Diffusion~\cite{Song21c} is a singularly powerful way to turn a random variable into a plausible density $\bx$. Furthermore, it can be guided so that $\bx$ obeys the constraint of~\eqref{eq:radon}, which makes it an effective tool for our purposes. Thus, this is what \ours{} relies on.

\parag{From Unguided to Guided Diffusion.}

The DDIM version of diffusion~\cite{Song21c} involves taking a sequence of $T$ denoising steps for $\{t_1 = 1 > t_2 > \ldots  > t_T =0 \}$ starting from $\bx_1$ being random noise and ending with $\bx_0$ being a plausible density distribution. At each iteration, $\bx_{t-1}$ is estimated from $\bx_{t}$ by computing
\begin{align}
\epsilon_{t|0}  & = \mN (\bx_t;t) \; , \nonumber \\
\bx_{t|0} & = (\bx_t - \beta_t \epsilon_{t|0}) / \alpha_t \; , \label{eq:DDPM} \\
\bx_{t-1}       & =  \alpha_{t-1} \bx_{t|0} + \beta_{t-1} \epsilon_{t|0} \; ,  \nonumber
\end{align}
where $\mN$ is a neural network trained to return an estimate of $\epsilon_{t|0}$ the noise yielding $\bx_t$ from $\bx_0$, $\bx_{t|0}$ is a plausible value of $\bx$ given the current noisy version $\bx_t$, and the $\alpha_t$ and $\beta_t$ are coefficients of the scheduler so that $\bx_{t-1}$ contains less noise than $\bx_t$, as presented in~\cite{Gao25a} whose notation we adopt in this paper.

In our case, we want to ensure that $\bx_0$ satisfies~\eqref{eq:radon} in addition to being a plausible density. An effective way is to enforce this constraint on successive estimates of $\bx_{t|0}$ in~\eqref{eq:DDPM} using a projector $P$ such that $\forall \bx$, $\bA P( \bx ) =   \by$. Each denoising iteration then becomes
\begin{align}
\epsilon_{t|0}  & = \mN (\bx_t;t) \; , \nonumber \\
\bx_{t|0} & = P((\bx_t - \beta_t \epsilon_{t|0}) / \alpha_t) \; , \label{eq:ProjectedDDPM} \\
\bx_{t-1}       & =  \alpha_{t-1} \bx_{t|0} + \beta_{t-1} \epsilon_{t|0} \; .  \nonumber
\end{align}
In practice, $P$ can be implemented using the pseudo inverse of $\bA$, as in~\cite{Wang22c}. In computed tomography, to our knowledge, the pseudo inverse of $\bA$ is rarely used directly for reconstruction and FBP or iterative methods are used instead~\cite{Liu23k, Chung23a, Xia23a}. In our work, the projector is implemented as successive gradient descent steps on  $\| \by - \bA \bx \| ^2$ iteratively decrementing $\bx$ by $\lambda d\bx$, where $d\bx = \bA^T (\by - \bA \bx)$ is the gradient of the square norm and $\lambda$ is a scalar that controls the step size.

\parag{Noise Estimation.}

A critical component of any DDIM algorithm is the network $\mN$ of~\eqref{eq:DDPM} and~\eqref{eq:ProjectedDDPM} that estimates the noise $\epsilon_{t|0}$. To this end, we designed a network architecture that treats depth slices as channels of a 2D U-Net with self-attention layers in the network to capture inter-slice dependencies~\cite{VonPlaten22}, as shown in Fig.~\ref{fig:full_pipeline}. This is a key design to make the ill-posed problem more tractable. To handle potentially arbitrary viewing angles, the network is structured so that the acquisition angle range $\theta$ and step $\Delta\theta$ are injected into the network bottleneck, after Fourier position encoding \cite{kishore25} and fed to a 2-layer MLP, shown in orange in Fig.~\ref{fig:full_pipeline}.

To further increase the predictive power of $\mN$, we modify its architecture so that it can take an additional input, in the form of additional slices, as shown in Fig.~\ref{fig:full_pipeline}. These slices are $C=\mR^{-1}(\by)$, the inverse Radon transform of $\by$ that can be acquired by FBP. Although $C$ is affected by the missing wedge problem, it still provides information that the network can exploit to generate a more meaningful noise estimation~\cite{Liu23k}. Besides, we train the diffusion model conditionally with the additional slices being $C$ and unconditionally with the slices being all zeros, which we denote as $\emptyset$, similarly to the classifier-free guidance (CFG)~\cite{Ho22a}. Thus we evaluate $\epsilon_{t|0}$ in~\eqref{eq:ProjectedDDPM}, we obtain an estimate with $\emptyset$ and another with $C$ and write
\begin{align}
\epsilon_{t|0}^{\emptyset}  &=  \mN ([\bx_t | \emptyset] ; t , \theta , \Delta\theta)  \; , \nonumber  \\
\epsilon_{t|0}^{C}               &=  \mN ([\bx_t | C] ; t , \theta , \Delta\theta)  \; ,  \label{eq:doubleEstimate} \\
\epsilon_{t|0}                      & = (1-s)\epsilon_{t|0}^{\emptyset}  + s\epsilon_{t|0}^{C} \; , \nonumber
\end{align}
where $[\cdot | \cdot]$ denotes the concatenation of the slices and $s$ is a scalar.

\parag{Handling Uncertainty.}

Equation~\eqref{eq:radon} is an idealization of the tomographic progress. In practice, tilt series always exhibit geometric distortions, even after careful calibration and alignment. This is particularly damaging when the number of images in the tilt series is small, such as when we consider an
$8^\circ$ and have only 5 views to work with, as the capability of fidelity term to averaging out misalignment via information of other views is decreased. 

To account for the distortions discussed above, we assign a per-voxel uncertainty measurement that reflects the consistency of voxel values obtained from different tilts. Specifically, for each tilt angle $i=1,\dots,T$, we project the tilt back to the 3D volume and get value $b_i(v)$ at voxel $v$. The uncertainty of voxel $v$ is then defined as the normalised variance of these values over all tilts:
\begin{equation}\label{eq:uncertainty}
\bu \;=\; \frac{ \frac{1}{T}\sum_{i=1}^T \big(b_i(v) - \bar b(v)\big)^2}{\mathrm{Var}_{\max}}
\end{equation}
Here, $\mathrm{Var}_{\max}=\frac{T+1}{4T}$ and normalisation ensure $u(v)\in[0,1]$. Voxels where different tilts agree yield low uncertainty and have higher weights, whereas voxels with inconsistent value from tilts exhibit high uncertainty and have lower weights. The guided reconstruction process is summarised in Algorithm~\ref{alg:sampling}.

\begin{algorithm}
\caption{Reconstruction steps}\label{alg:sampling}
\begin{algorithmic}[1]
\Require diffusion steps $N$, measurement $\by$, angle range $\theta$, angle step $\Delta\theta$, cfg scale $s$\\

$\bx_N \sim \bN(0,\bI)$, $\bC = \mR^{-1}(\by)$, $\bu$.

\For{$t=N, N-1, \ldots, 1$}
    \State $\epsilon_{t|0}^{\emptyset}  =  \mN ([\bx_t | \emptyset] ; t , \theta , \Delta\theta) $
    \State $\epsilon_{t|0}^{C} = \mN ([\bx_t | C] ; t , \theta , \Delta\theta) $
    \State $\epsilon_{t|0} = (1-s)\epsilon_{t|0}^{\emptyset}  + s\epsilon_{t|0}^{C} $
    \State $\bx_{t|0} = (\bx_t - \beta_t \epsilon_{t|0}) / \alpha_t$
    \State $\tilde{\bx}_{t|0} = P(\bx_{t|0})$
    \State $\bx_{t-1} = \alpha_{t-1} \left ( \bu \bx_{t|0} + (1-\bu)\tilde{\bx}_{t|0} \right ) + \beta_{t-1} \epsilon_{t|0} $
\EndFor \\
\Return $\bx_0$
\end{algorithmic}
\end{algorithm}

\subsection{Training the Network using FIB-SEM Data}
\label{sec: simulator}

Training the noise prediction network $\mN$ of~\eqref{eq:doubleEstimate} requires ground truth data. However, reconstructions from real STEM tilting series are all affected by the missing wedge problem, making artefact-free ground truth data essentially unattainable. To overcome this limitation, we treat the FIB-SEM signal as an estimate of the heavy atom density and use it to synthesise STEM series in high-angle angular dark field (HAADF) mode. The density volumes then serve as a ground-truth substitute for training purposes. In the following, we describe the relationship between these densities and the pixel values in the synthetic HAADF projections.

\parag{FIB-SEM Contrast Model}

Let $S(p)$  denote the intensity of a  FIB-SEM voxel at position $p=(x,y,z)$. We {approximate} its dependence on the local heavy-atom content $Z(p)$ as
\begin{equation}\label{eq:fibsem}
S(p) = a Z(p)^{\mu} \; ,
\end{equation}
where $a$ and $\mu$ are parameters depending {by the imaging conditions}.

\parag{HAADF contrast model}
For thin specimens, the log-ratio {method relates} the thickness $t$ to the inelastic mean free path $\lambda$ by
\begin{equation}\label{eq:logratio}
\ln\left( \frac{I_0}{I} \right) = \frac{t}{\lambda} \; ,
\end{equation}
where $I_0$ is the total area under the EELS spectrum and $I$ is the zero-loss intensity.

{A simplified expression for} $\lambda$ is
\begin{equation}\label{eq:imfp}
\lambda = \frac{E_0}{Z^{\,r} \cdot \rho} \; ,
\end{equation}
{ with} $E_0$ {the energy beam}, $Z$ the atom{ic} number, $\rho$ {the sample} density~\cite{Egerton11}. Combining~\eqref{eq:logratio} and~\eqref{eq:imfp} gives
\begin{equation}\label{eq:temexp}
I = I_0 e^{-C \cdot Z^r \cdot t} \; ,
\end{equation}
where $C$ is a {constant that depends on} beam energy and material properties. For thicker specimens, the sample can be approximated as a stack of thin slices. In this case, the total attenuation is obtained by integrating the contribution of each slice along the beam path. This line integration being mathematically equivalent to the Radon transform, denoted by $R(\cdot)$, we write
\begin{equation}\label{eq:tem_beerlambert}
I = I_{0} e^{-C \cdot R(Z^r)} \; ,
\end{equation}

Since the HAADF contrast is approximately proportional to the inelastic component,  $I_0 - I$, we {obtain}
\begin{equation}\label{eq:I_haadf}
I_{\mathrm{HAADF}} = k*(1 - e^{-C \cdot R(Z^r)}) \; .
\end{equation}
{with proportionality constant $k$.}

\parag{FIB-SEM to HAADF Mapping}
Injecting the heavy atom content $Z$ of~\eqref{eq:fibsem}  into~\eqref{eq:I_haadf} yields the desired mapping from FIB-SEM to {synthetic HAADF intensity}
\begin{equation}\label{eq:sem2tem}
I_{\mathrm{HAADF}} \approx k* \left ( 1 - e^{-C \cdot R({S}^{\gamma})} \right ) \; ,
\end{equation}
where $C$ and $\gamma$ characterize the imaging configuration and sample properties. During training, tilt angle $\theta$ and increments $\Delta\theta$ are randomly sampled from predefined ranges and Equation ~\ref{eq:sem2tem} is used to generate STEM tilt series used for training.

\rev{In practice, the simulator described above provides an approximate mapping between FIB–SEM volumes and HAADF-STEM projections and therefore cannot fully capture all experimental variations. In particular, factors such as beam intensity, beam energy, specimen composition, and detector response may influence the measured contrast and noise characteristics. Variations in these parameters can lead to discrepancies between the simulated projections used during training and the projections acquired in real experiments. To reduce this gap, the parameters of Eq. (12) are chosen such that the statistics of the simulated tilts closely match those of real STEM images. Besides, the simulator can be used for data augmentation over a broad range of parameter combinations, thereby improving the model’s robustness to different conditions. In addition, the reconstruction procedure incorporates a projection consistency step at each diffusion iteration, which constrains the reconstructed volume to remain consistent with the measured projections. This data consistency enforcement helps mitigate moderate mismatches between the simulator and the real imaging conditions. Nevertheless, the simulator remains an approximation of the underlying imaging physics, and large deviations in imaging parameters or specimen composition may still affect the reconstruction accuracy. Exploring more accurate simulation models and domain adaptation strategies is therefore an important direction for future work.} 

%% file: tex/material.tex

\section{Materials and Experimental Setup} \label{sec:material_and_method}

\rev{In this work, we use brain and HeLa cell samples in  their resin blocks and were prepared following the protocol of Knott \textit{et al.} }\revshort{~\cite{Knott08a}.}\rev{ The biological samples and resin block preparations were produced at the BioEM facility at EPFL. Thin sections of approximately 300, 500, and 1000~nm were prepared for STEM experiments, while sections of approximately 300~nm were prepared for TEM experiments using an ultramicrotome (Leica EM UC7) equipped with a diamond knife (Diatome$^{\text{\textregistered}}$).
These biological samples were then imaged using either FIB-SEM or TEM/STEM microscopes as described below.}

\input{fig/dataset_sample}

\subsection{Acquiring FIB-SEM volumes}
\label{sec:fibsem}

\rev{Four FIB-SEM volumes were acquired and aligned according to the protocol of Knott \textit{et al.}}\revshort{~\cite{Knott08a}.}\rev{ Three volumes contain mitochondria and one contains synapses, from two types of biological samples, namely brain tissue and HeLa }\revshort{cells~\cite{Knott08a,Knott11,Riguet21}.}\rev{ Examples are shown in}\revshort{ Fig.~\ref{fig: dataset_samples}.}

\rev{Each volume was split into training, evaluation, and testing sub-volumes with a ratio of 6:2:2. From each sub-volume, we extracted patches of size $128\times128\times40$ that contain at least one target structure. The in-plane size of $128\times128$ is sufficient to cover a single mitochondrion or synapse in these FIB-SEM data, while the depth of 40 slices approximates the physical section thickness used in STEM experiments on biological samples.}

\rev{To increase the diversity of the training data, we applied 3D augmentations including flipping, resizing, affine, elastic, and perspective transformations. We simulated tilt series with angular ranges of $10^\circ$, $30^\circ$, and $60^\circ$, each symmetric around $0^\circ$ and sampled at $1^\circ$ increments. Examples are shown in }\revshort{Fig.~\ref{fig:sem_to_tem}.}

\subsection{Acquiring (S)TEM Tilt Series}
\label{sec:stem}

\rev{We additionally constructed datasets directly from real tilt series. Six real tilt series were used: three STEM tilt series of single-cell mitochondria and three TEM tilt series of brain-cell synapses. Each tilt series contains 41 or 61 projections spanning a total angular range of $80^\circ$ or $120^\circ$ with $2^\circ$ increments, acquired at a resolution of $1024\times1024$.}

\rev{For TEM tomography, digital tilt-series images were acquired automatically using a Tecnai F20 electron microscope (ThermoFisher Scientific$^{tm}$) operated at 200~kV and equipped with a Falcon K3 camera, using the ThermoFisher Tomography TEM software, with a pixel size of 0.415~nm. For STEM tomography, digital tilt-series images were acquired automatically using a Titan Themis microscope (ThermoFisher Scientific$^{tm}$) operated in STEM mode at 300~kV and equipped with a Fischione photomultiplier (PMP) HAADF detector, also using the ThermoFisher Tomography STEM software with a pixel size of 2.29~nm.}

\rev{For both TEM and STEM tomography, 3D reference volumes were reconstructed using }\revshort{Aretomo~\cite{Zheng22a}.}\rev{ The resulting volumes were then cropped to the regions containing the biological structures of interest before further analysis. Because these real tilt series are still limited-angle acquisitions, the reconstructed volumes remain affected by missing-wedge artifacts. We therefore retained only regions with relatively few artifacts for evaluation. For each target structure, the corresponding reconstructed sub-volume was cropped and down-sampled to $128\times128\times40$ to serve as the reference volume.}

\rev{Due to the limited availability of real tilt series, these data were used exclusively for testing. All models were trained on the simulated datasets derived from FIB-SEM volumes described above.}

%% file: fig/dataset_sample.tex
\begin{figure*}
  \centering
  \begin{subfigure}[b]{0.23\textwidth}
    \centering
    \includegraphics[width=\linewidth]{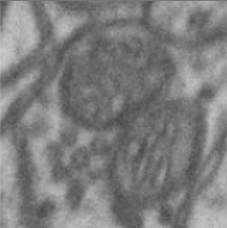}
    \subcaption{}\label{fig:a}
  \end{subfigure}\hfill
  \begin{subfigure}[b]{0.23\textwidth}
    \centering
    \includegraphics[width=\linewidth]{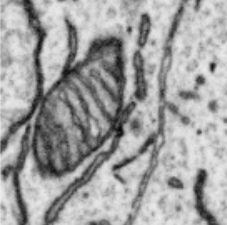}
    \subcaption{}\label{fig:b}
  \end{subfigure}\hfill
  \begin{subfigure}[b]{0.23\textwidth}
    \centering
    \includegraphics[width=\linewidth]{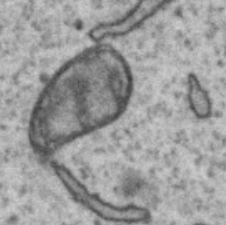}
    \subcaption{}\label{fig:c}
  \end{subfigure}\hfill
  \begin{subfigure}[b]{0.23\textwidth}
    \centering
    \includegraphics[width=\linewidth]{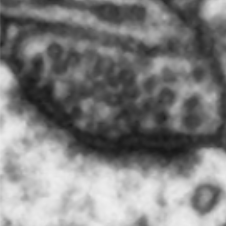}
    \subcaption{}\label{fig:d}
  \end{subfigure}
  \begin{subfigure}[b]{0.23\textwidth}
    \centering
    \includegraphics[width=\linewidth]{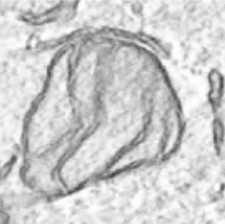}
    \subcaption{}\label{fig:e}
  \end{subfigure}\hfill
  \begin{subfigure}[b]{0.23\textwidth}
    \centering
    \includegraphics[width=\linewidth]{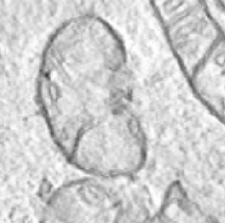}
    \subcaption{}\label{fig:f}
  \end{subfigure}\hfill
  \begin{subfigure}[b]{0.23\textwidth}
    \centering
    \includegraphics[width=\linewidth]{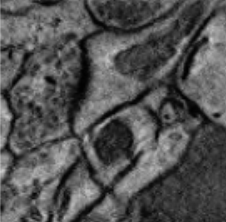}
    \subcaption{}\label{fig:g}
  \end{subfigure}\hfill
  \begin{subfigure}[b]{0.23\textwidth}
    \centering
    \includegraphics[width=\linewidth]{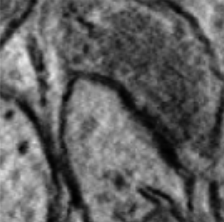}
    \subcaption{}\label{fig:h}
  \end{subfigure}

  \caption{Samples of different datasets. FIB-SEM image of (a) mitochondria in brain cells;
  (b) and (c) mitochondria in two different HeLa cells; (d) synapse in brain cells.
  Reconstruction using AreTomo from real TEM tilts of (e) and (f) mitochondria,
  (g) and (h) synapse}
  \label{fig: dataset_samples}
\end{figure*}

%% file: tex/results.tex

\section{Results and Discussion}

We compare \ours{} with several baselines on both real and synthetic data. \rev{The synthetic data comprises the tilt series and patches from the FIB-SEM data of}\revshort{ Section~\ref{sec:fibsem},}\rev{ which are also used both for training purposes and qualitative evaluation. The real data consists of the STEM tilt series of}\revshort{ Section~\ref{sec:stem}}\rev{, on which we run our algorithm }\revshort{{\it without}}\rev{ further re-training. Owing to the missing wedge problem, true ground-truth volumes are not available for such data. We therefore perform a limited amount of quantitative evaluation on it and a more thorough one on the synthetic data, along with an ablation study.} 

\subsection{Implementation Details}

\paragraph{Training Procedure}
\rev{Our training procedure follows the}\revshort{ Diffusers~\cite{VonPlaten22}}\rev{ framework for unconditional image generation. The model is trained with the Adam optimizer using a cosine warm-up schedule. The learning rate is set to $1\times10^{-4}$ with 500 warm-up steps. Exponential Moving Average (EMA) is used to stabilize training. We train the model for 2,000 epochs with a batch size of 32.}

\paragraph{Reconstruction Parameters}
\rev{Simulator parameters are estimated by minimizing the histogram and style difference between synthesized projections and real tilts which have similar thickness (e.g. 300 nm). We conduct coarse grid search on validation set to find the combination of gradient step size $\lambda$ and guidance scale $s$. The parameters are set to $\left( \lambda, s\right) = \left(10, 2\right)$. The number of gradient steps of every projection iteration in reverse diffusion step is set to 1.}

\begin{figure*}[!t]
    \centering
    \revframe{%
        \includegraphics[width=0.6\textwidth]{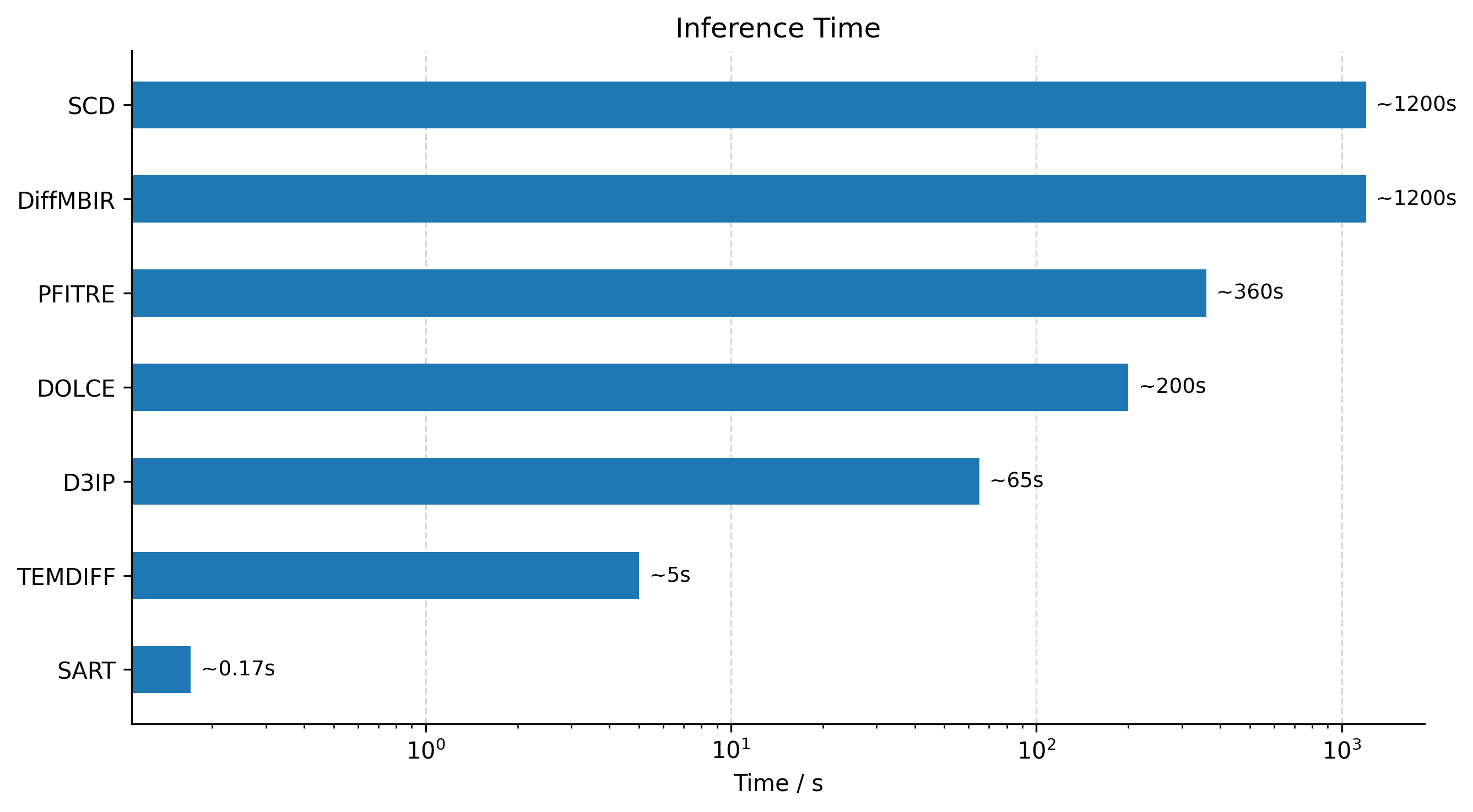}
    }
    \caption{\rev{Time of different methods to reconstruct a volume of $40\times128\times128$ with angular range of $10^\circ$ and angular step of $1^\circ$.}
    }
\label{fig: speed_comparison}
\end{figure*}

\paragraph{Computation Cost}
\rev{All experiments were conducted on a server equipped with an NVIDIA A100 GPU with 80 GB of memory. Training }\revshort{\ours{}}\rev{ with a combination of several angular ranges and step sizes, for example, $\theta \in (8^\circ, 10^\circ, 12^\circ, 14^\circ)$ and $\Delta\theta \in (1^\circ, 2^\circ, 3^\circ)$, takes approximately 2.5 days on a V100 GPU (32 GB) and about 1 day on an A100 GPU (80 GB). In general, the broader the range of angles and step sizes used for training, the longer the training time. Reconstructing an 11-tilt sample of size $128 \times 128$ with 50 steps takes about 5 seconds. The inference times of different methods are compared and visualized in}\revshort{ Fig.~\ref{fig: speed_comparison}}\rev{. Although SART is the fastest method, }\revterm{\ours{}}\rev{ achieves a good balance between computational efficiency and reconstruction performance, offering relatively short inference time while delivering leading results.}

\subsection{Simulator}
\rev{We use the simulator described in}\revshort{ Section~\ref{sec: simulator}}\rev{ to transfer synthesized tilts to the STEM HAADF style, as shown in}\revshort{ Fig.~\ref{fig:sem_to_tem}.}\rev{ The parameters $C$ and $\gamma$ in}\revshort{ Eq.~\eqref{eq:sem2tem}}\rev{ are adjusted so that the statistics of the simulated HAADF images more closely match those of real experimental data than those obtained by direct Radon transform, as illustrated by the histogram comparison in}\revshort{ Fig.~\ref{fig:sem_to_tem}(d).}

\revshort{Figure~\ref{fig:sem_to_tem}(e)}\rev{ further provides a feature-level (latent) distance comparison where the latent distance is measured in the feature space defined in}\revshort{ \cite{Gatys16a}.}\rev{ We define the relative latent distance as the ratio between (i) the average distance from the source images to the real tilt images and (ii) the average pairwise distance among the real tilt images themselves, where the source images are either simulated tilts or tilts obtained by direct Radon transform. The figure shows that the relative latent distance decreases as $C$ and $\gamma$ are optimized. After about 10 adjustment steps, the simulated tilts are already closer to the real tilts in latent space than the direct-Radon-transform baseline, and the distance continues to decrease with additional steps. The minimum relative latent distance is approximately 7.0 after 200 steps (the simu\_tilts in the figure), and it can be further reduced to about 6.8 when style loss is incorporated into the adjustment process (the simu\_tilts\_combi in the figure).}

\input{fig/physical_model}

\subsection{Baselines and Evaluation Metrics}

To comprehensively evaluate \ours{}, seven baselines spanning iterative method, convolution neural network, diffusion model, and hybrid strategies are compared: SART in the Aretomo, a classic iterative method commonly used in biological tomography; \pfitre{}, an ADMM-based method with U-Net-based artifact suppression; \dolc{}, a conditional 2D diffusion approach; \mbir{}, an MBIR-style method with vertical consistency regularization and an unconditional 2D diffusion prior; \scd{} and \ddip{}, LoRA-based~\cite{Hu22} test-time finetuning on pre-trained 2D diffusion models for 3D reconstruction.

We use the official codebases provided by the authors with minimal modifications to support our setup, preserving each method's original structure. Hyper parameters are first selected according to the guidelines in the published articles, then further tuned via grid search on a subset of the validation set. All methods were trained until convergence.

\begin{figure*}[!t]
  \centering

  \begin{subfigure}[b]{0.22\textwidth}
    \centering
    \includegraphics[width=\linewidth]{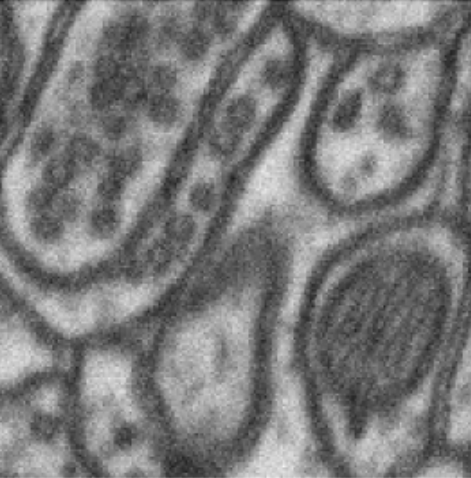} 
    \subcaption{}\label{fig:recon_a}
  \end{subfigure}\hspace{0.01\textwidth}
  \begin{subfigure}[b]{0.22\textwidth}
    \centering
    \includegraphics[width=\linewidth]{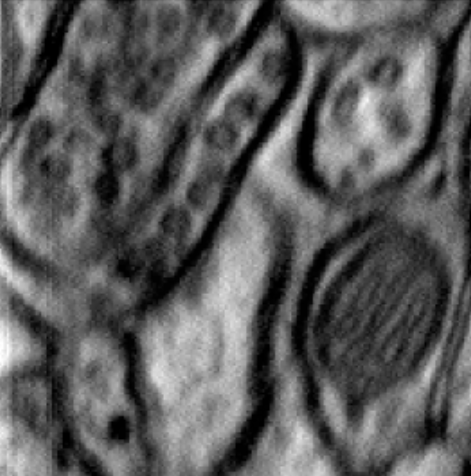} 
    \subcaption{}\label{fig:recon_b}
  \end{subfigure}\hspace{0.01\textwidth}
  \begin{subfigure}[b]{0.22\textwidth}
    \centering
    \includegraphics[width=\linewidth]{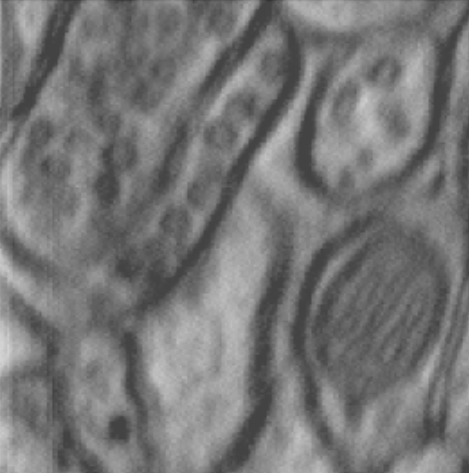} 
    \subcaption{}\label{fig:recon_c}
  \end{subfigure}

  \caption{Example of reconstruction before and after histogram adjustment.
  (a) ground truth. (b) reconstruction from Aretomo before adjustment with lower PSNR.
  (c) after adjustment with higher PSNR.}
  \label{fig:histo_adjust}
\end{figure*}

For quantitative evaluation, we use Peak Signal-to-Noise Ratio (PSNR) and Structural Similarity Index Measure (SSIM) \rev{between reconstructed volumes and reference volumes to assess reconstruction quality. We also report the root-mean-square error between the projections of the reconstructed volumes and the corresponding tilt images as a measure of residual fidelity error (RFE). When ground-truth volumes are available, RFE is calculated from tilt images within the angular range used for reconstruction, as shown in}\revshort{ Table~\ref{tabel: quantitative_results_all}.}\rev{ When they unavailable, RFE is calculated instead from tilt images outside the angular range used for reconstruction, since it offers a more meaningful assessment of whether the reconstruction is consistent with unseen tilt images, as shown in}\revshort{ Table~\ref{tabel: quantitative_results_real}.}\rev{ In addition, to quantify 3D structural consistency, we compute the 3D perceptual loss (PL) between features extracted using}\revshort{ Med3D~\cite{Chen19d} in MONAI~\cite{Ca22}}\rev{ from reconstructed volumes and reference volumes.} Besides, limited-angle reconstructions can exhibit global grayscale shifts as shown in Fig.~\ref{fig:histo_adjust}, to mitigate its impact on the metrics, we apply a linear histogram matching to align the first (Q1) and third (Q3) quartiles between the reconstructed and ground-truth volumes before metric computation.

\subsection{Performance on Real TEM tilts}
\begin{figure*}[!t]
    \centering
    \includegraphics[width=1\textwidth]{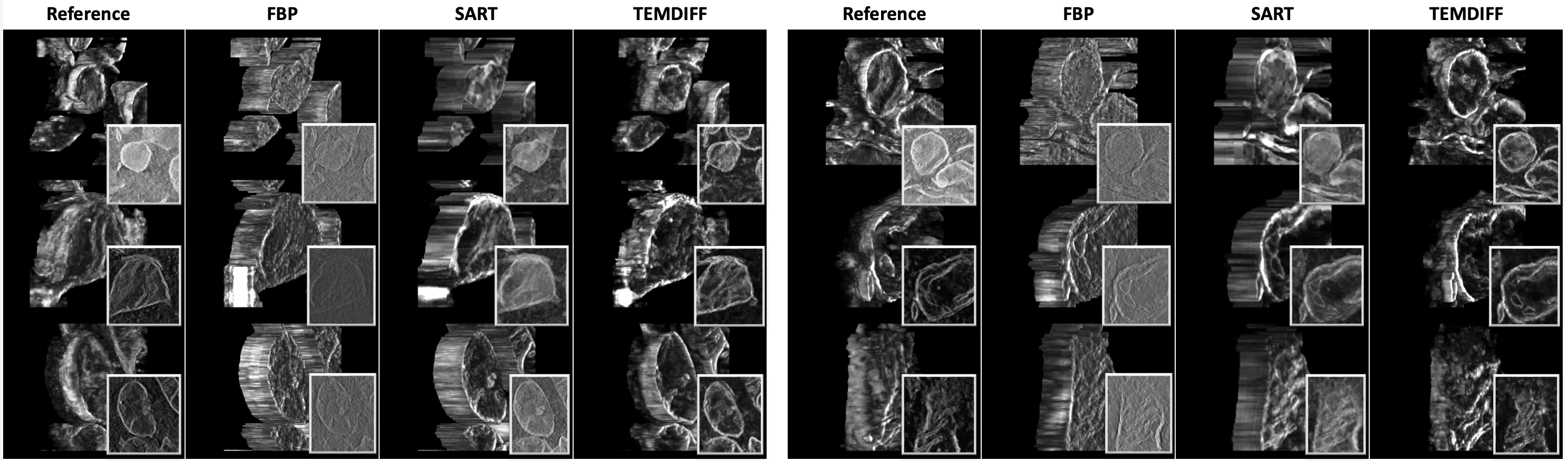}
    \caption{Reconstructions of FBP, SART and \ours{} with real tilts of $8^\circ$ angular range and $1^\circ$ or $2^\circ$ increments. The reference volumes are reconstructed using AreTomo with all available tilts ($80^\circ$ or $120^\circ$). Each main figure is the 3D view of reconstruction and its bottom-right inset corresponds to one slice of 2D view. \ours{} produces more realistic and clearer reconstructions with good contrast on both 3D and 2D views. In 3D views, the contrast of FBP and SART are manually enhanced for better visual quality.
    }
\label{fig: error_comparison_real_8}
\end{figure*}

The entire \ours{} pipeline is applied to raw tilt series of biological samples with different thicknesses ($\approx$300nm, 500nm, 1000nm), acquired over ranges of $8^\circ$ with $1^\circ$ or $2^\circ$ increments using either TEM or STEM. Fig.~\ref{fig: error_comparison_real_8} depicts the results.  Without any retraining or fine-tuning, our method yields clear and plausible reconstructions with enhanced contrast and structural coherence, showing qualitatively better results than those of FBP and SART, with fewer artifacts and increased clarity. For comparison purposes, reference reconstructions were produced with AreTomo using larger tilt ranges covering $120^\circ$ or $80^\circ$. In some cases, as in the top row of Fig.~\ref{fig: error_comparison_real_8}, large-angle reconstructions exhibit blurred membranes whereas \ours{} reduces or even suppresses these artifacts and produces structures with sharper boundaries and better visual contrast.

\subsection{Comparative Results on Synthetic Data}
\input{fig/results_sem}

\subsubsection{Datasets from FIB-SEM}
\begin{figure*}[!t]
    \centering
    \revframe{%
        \includegraphics[width=\textwidth]{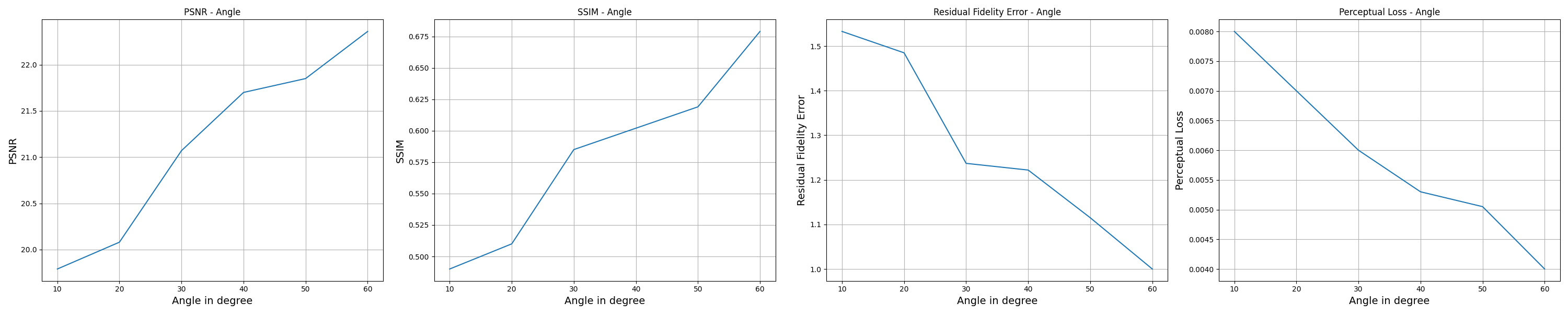}
    }
    \caption{\revshort{\ours{} performance with different visible angular range.}
    }
\label{fig: more_angle}
\end{figure*}

Each method is trained and evaluated on simulated FIB-SEM datasets with angular coverages of $10^\circ$, $30^\circ$ and $60^\circ$. Quantitative results appear in Table~\ref{tabel: quantitative_results_all}. Consistently, \ours{} achieves either the best or the second best values for SSIM\rev{, RFE and PL}. The advantage is particularly pronounced in the extreme $10^\circ$ ranges. On the Hela 1 dataset, \pfitre{} is slightly better than \ours{}. On the synapse dataset, \mbir{} and \ddip{} yields better PSNR, but \ours{} achieves higher SSIM \rev{and similar or higher RFE and PL}, indicating more accurate structural recovery. Across mitochondria and synapses volumes from difference biological samples, \ours{} delivers the most consistent improvements, especially at $10^\circ$, where enforcing 3D coherence is critical. \revshort{Evaluations of \ours{}}\rev{ covering $\left(10^\circ, 20^\circ, 30^\circ, 40^\circ, 50^\circ, 60^\circ\right)$ on the Brain dataset are also provided in}\revshort{ Fig.~\ref{fig: more_angle}}\rev{ to show the quality trend over many different ranges.}

\input{fig/results_simtem}

\subsubsection{Datasets from Real Tilts}
\begin{figure*}
    \centering
    \revframe{%
        \includegraphics[width=1\textwidth]{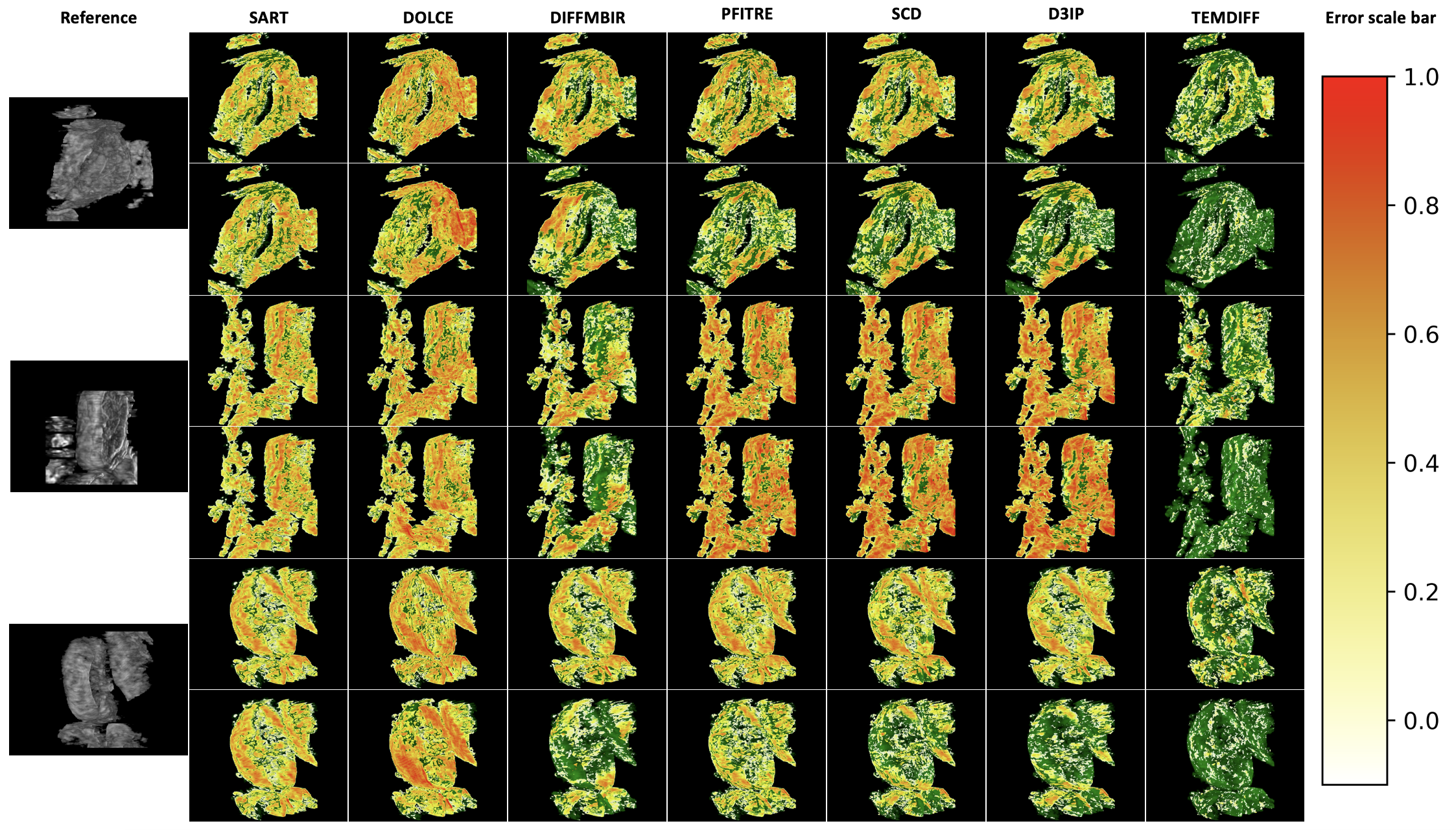}
    }
    \caption{Errors of three mitochondria reconstructed by different methods with an angular range of $10^\circ$ (odd rows) and $60^\circ$ (even rows). Reference reconstructions are obtained via Aretomo with tilting series covering $120^\circ$ (the first mitochondrion) and $80^\circ$ (the second and third mitochondria). For clarity, only the errors on membrane are plotted.
    Yellow and red show detailed error distribution, i.e. yellow means low error while red indicates high errors. \ours{} produces less error comparing with other methods.
    }
\label{fig: error_comparison_mito_allinone}
\end{figure*}

\begin{figure*}
    \centering
    \revframe{%
        \includegraphics[width=1\textwidth]{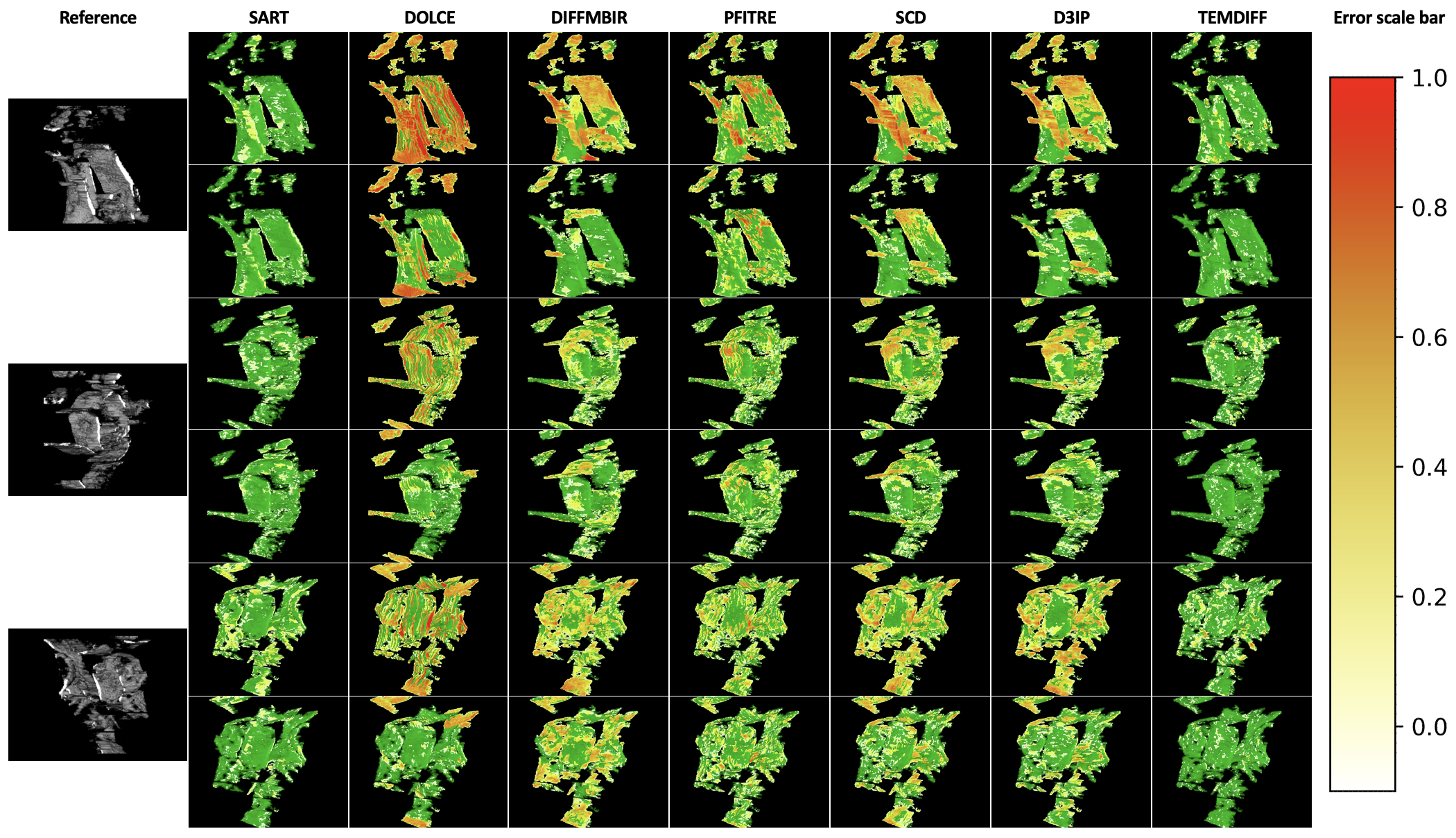}
    }
    \caption{
    Errors of three synapses reconstructed by different methods with angle range of $10^\circ$ (odd rows) and $60^\circ$ (even rows). Reference reconstructions are obtained via Aretomo with tilting series covering $120^\circ$ (the first synapse) and others with $80^\circ$ (the second and third synapses). For clarity, only the errors on membrane are plotted.
    Yellow and red show detailed error distribution, i.e. yellow means low error while red indicates high errors. \ours{} produces less error comparing with other methods.
    }
\label{fig: error_comparison_syna_allinone}
\end{figure*}

Because the reference reconstructions from AreTomo (80$^\circ$--120$^\circ$) still contain missing-wedge artifacts, quantitative metrics must be interpreted carefully. In particular, a high SSIM may reflect either faithful structure reconstruction or the reproduction of artifacts. \rev{We therefore use tilt images as ground truth and report RFE calculated from tilt images outside the angular range used for reconstruction, since this offers a more meaningful assessment of whether the reconstruction is consistent with unseen real tilt images. In addition, PSNR, SSIM, and PL should be interpreted together and considered alongside qualitative visualizations }\revshort{(Table~\ref{tabel: quantitative_results_real}; }\revshort{Figs.~\ref{fig: error_comparison_mito_allinone}--\ref{fig: error_comparison_syna_allinone}).}

On mitochondria data, \ours{} yields the highest PSNR, SSIM\rev{, RFE and PL} for both $10^\circ$ and $60^\circ$ ranges. On synapse data, at 10°, SART attains the highest PSNR but fails to maintain structural quality, resulting in reduced SSIM\rev{, RFE and PL}.

\begin{figure*}[!t]
    \centering
    \revframe{%
        \includegraphics[width=1\textwidth]{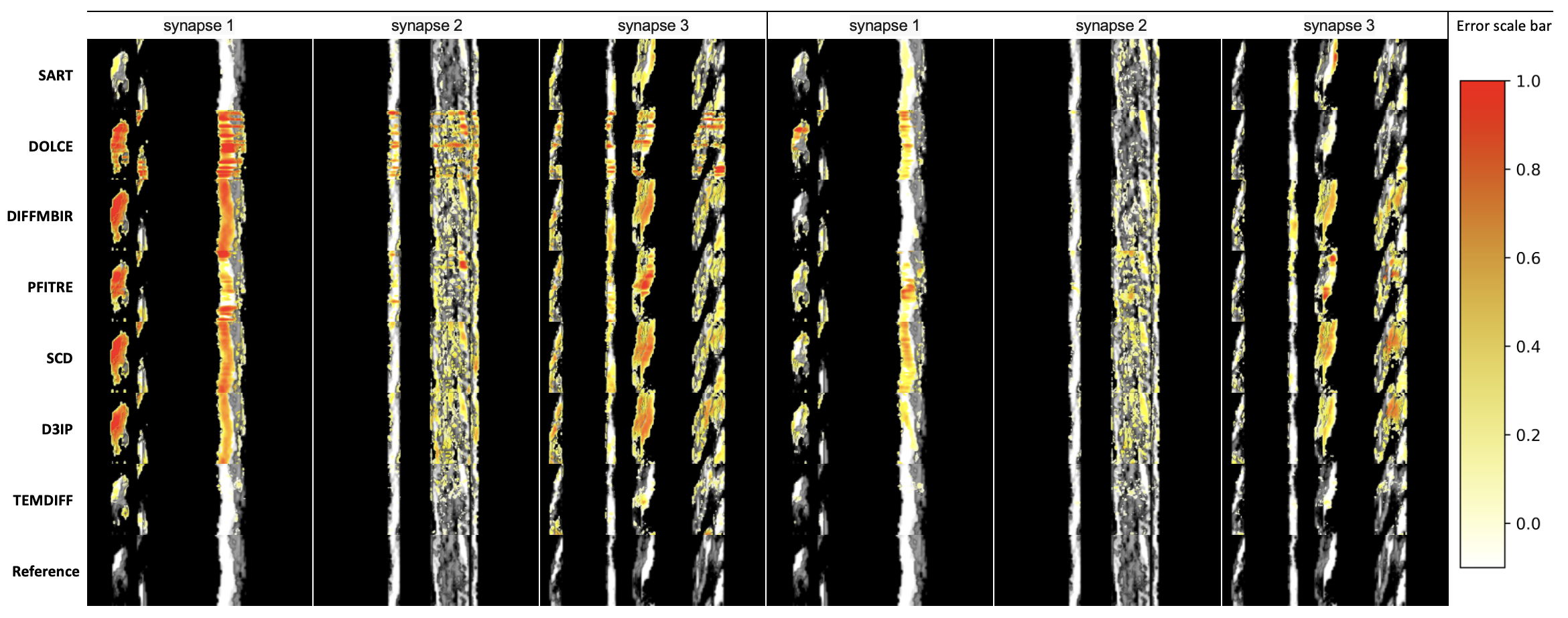}
    }
    \caption{
     Errors on vertical cross-section along x axis of synapses reconstructed from different methods. 
     Yellow and red show detailed error distribution. 
     For clarity, only the errors on membrane are plotted.
     In both $10^\circ$ (left) and $60^\circ$ (right) cases, \ours{} produces more plausible reconstructions with less errors than other methods.
    }
\label{fig: vertical_comparison}
\end{figure*}

Comparing the error map on vertical cross-section along x axis of reconstructions in Fig.~\ref{fig: vertical_comparison}, it shows that at $10^\circ$, \dolc{}, \pfitre{}, and \scd{} exhibits inter-slice inconsistency (horizontal banding artifacts), and \mbir{} and \ddip{} misses fine details despite its explicit Z-TV prior. Since the error map is calculated on the membrane shape which neglects the errors costed by streaking, SART results show less errors without achieving a good reconstruction. 
At $60^\circ$, all methods recover better structures while still suffering from the same limitation as $10^\circ$. In contrast, \ours{} suppresses streaking and yields good cross-slice consistency. Visual inspection (Figs.~\ref{fig: error_comparison_mito_allinone}-\ref{fig: error_comparison_syna_allinone}) further confirms that \ours{} produces the most consistent 3D morphology across slices. 
 
Overall, when judged by both quantitative metrics and qualitative volume renderings, \ours{} achieves the best balance between structure preservation and artifact suppression on our tests, particularly on the narrow end of the angular range.

\subsection{Ablation Study}

\paragraph{Projector and CFG}
\rev{To assess the contributions of projector guidance and classifier-free guidance, we compare the reconstruction performance of}\revshort{ \ours{}}\rev{ with and without these two components on the validation set of the Brain dataset at a $10^\circ$ angular range and a $1^\circ$ angular step. When the two components are enabled, we fix both weights $\lambda=1$ and $s=1$. As reported in}\revshort{ Table~\ref{table: ablation_guidance},}\rev{ both components consistently enhance reconstruction quality across all the four metrics.}
\begin{table}
\caption{\rev{Performance comparison of different components.}}
\label{table: ablation_guidance}
\centering
\small
\revframe{%
    \begin{tabular}{cccccc}
    \toprule
    \multirow{2}{*}{Projector} & \multirow{2}{*}{CFG} & \multicolumn{4}{c}{Metrics} \\
    \cmidrule(lr){3-6}
    & & PSNR$\uparrow$ & SSIM$\uparrow$ & RFE$\downarrow$ & PL$\downarrow$ \\
    \midrule
    \multirow{2}{*}{w/o} & w/o & 16.15 & 0.199 & 6.570 & 0.044 \\
                       & w/ & 18.92 & 0.367 & 2.460 & 0.017 \\
    \midrule
    \multirow{2}{*}{w/} & w/o & 18.32 & 0.204 & 5.939 & 0.038 \\
                       & w/ & 19.06 & 0.374 & 2.277 & 0.015 \\
    \bottomrule
    \end{tabular}
}
\end{table}

\begin{figure*}[!t]
    \centering
    \revframe{%
        \includegraphics[width=0.8\textwidth]{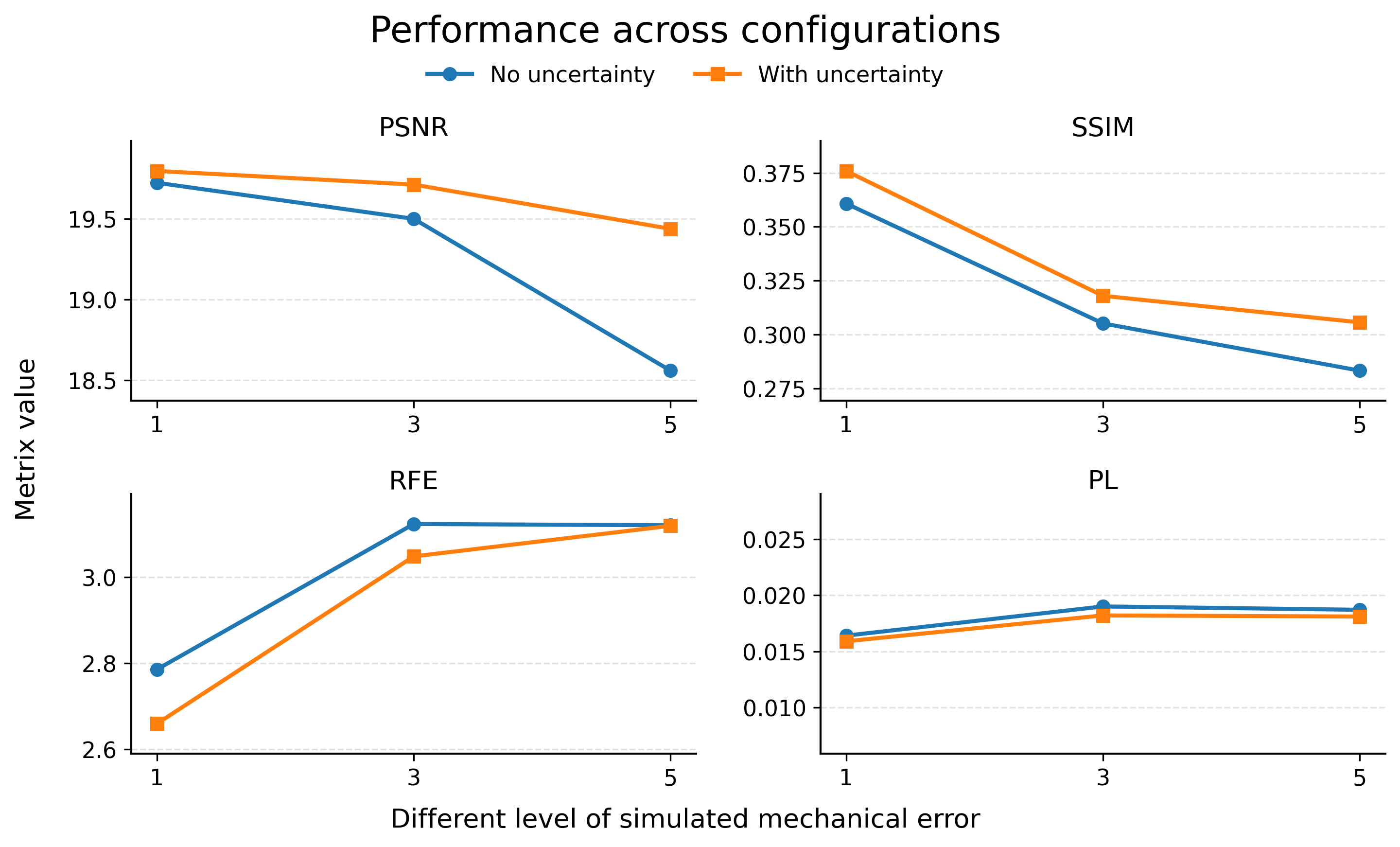}
    }
    \caption{\rev{Performance variation of pipelines with and without uncertainty term under 3 different mechanical error levels. High PSNR or SSIM, and low RFE or PL indicate better quality.}}
\label{fig: uncertainty}
\end{figure*}

\paragraph{Uncertainty term}
\rev{We evaluate the effectiveness of the uncertainty term using a pretrained model applied to synthesized tilt series from the Brain validation dataset, at a $10^\circ$ angular range and a $1^\circ$ angular step. The synthesized data include simulated mechanical errors, specifically: inter-tilt misalignment (in pixels), tilt-axis shift in the X--Z plane (in pixels), and tilt-axis rotation in the X--Y plane (in degrees). We consider three error levels: 1, 3, and 5 pixels for misalignment and axis shift, and 1, 3, and 5 degrees for axis rotation. For each level, the error values are sampled uniformly within the corresponding range. We compare the PSNR, SSIM, RFE, and PL of pipelines with and without the uncertainty term, as summarized in}\revshort{ Fig.~\ref{fig: uncertainty}.}

\begin{figure*}[!t]
    \centering
    \revframe{%
        \includegraphics[width=0.7\textwidth]{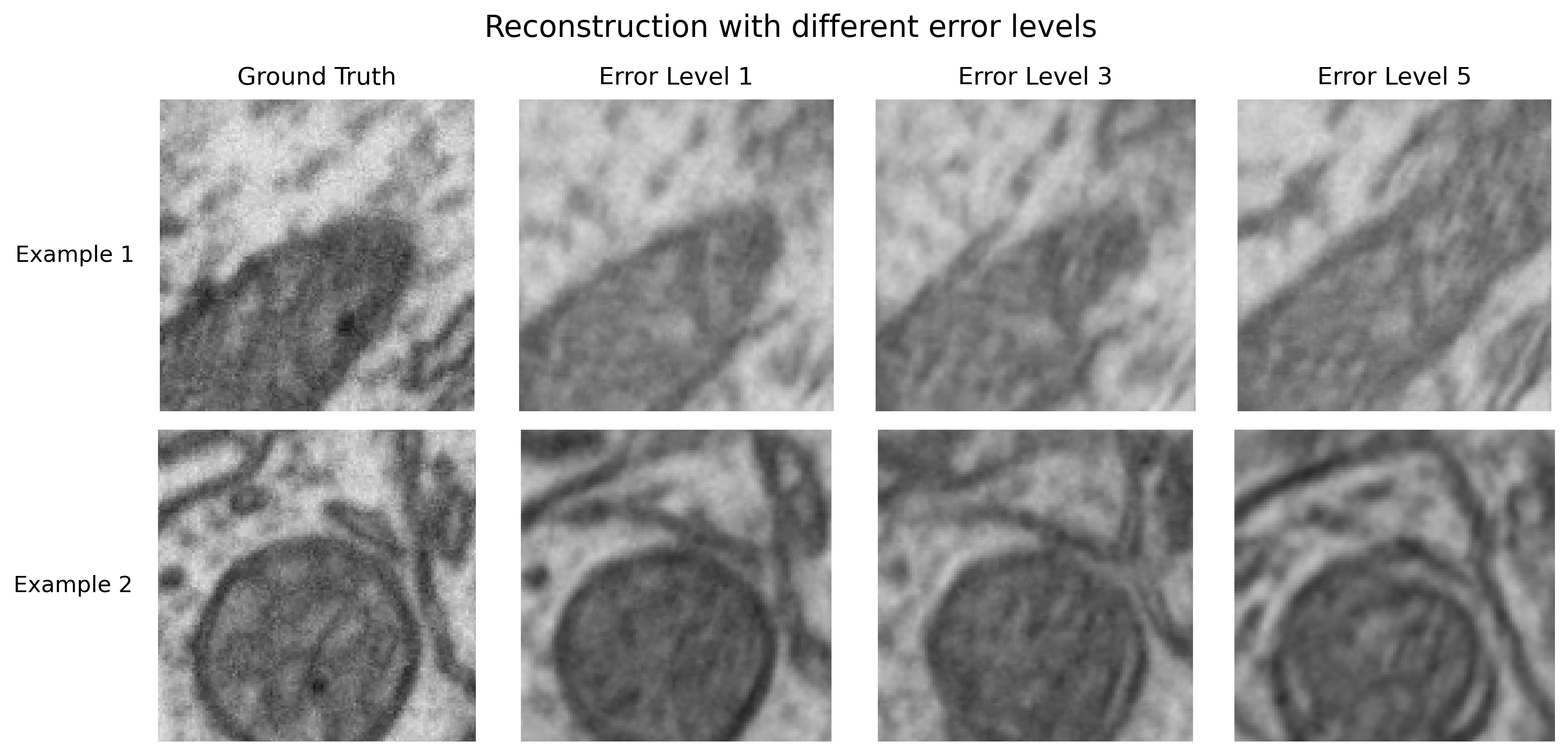}
    }
    \caption{\rev{Examples of reconstructions at different error levels. From level 3 onward, the major structural features of the target shapes are reconstructed incorrectly, indicating that the uncertainty weighting is insufficient to fully compensate for geometric errors start from this magnitude.}
    }
\label{fig: uncertainty_cases}
\end{figure*}

\rev{As the error level increases, the performance of both pipelines degrades, as expected, since the uncertainty term is designed to mitigate the effect of mechanical errors rather than eliminate it. Nevertheless, the pipeline with the uncertainty term consistently outperforms the version without it in all the four metrics across all tested error levels, indicating improved robustness to mechanical perturbations. These results confirm the benefit of incorporating the uncertainty term. To further illustrate this behavior, we visualize reconstructed samples at different error levels in}\revshort{ Fig.~\ref{fig: uncertainty_cases},}\rev{ highlighting when the uncertainty strategy begins to fail to preserve the major structural features of the target shapes.}

\begin{figure*}[!t]
    \centering
    \revframe{%
        \includegraphics[width=0.7\textwidth]{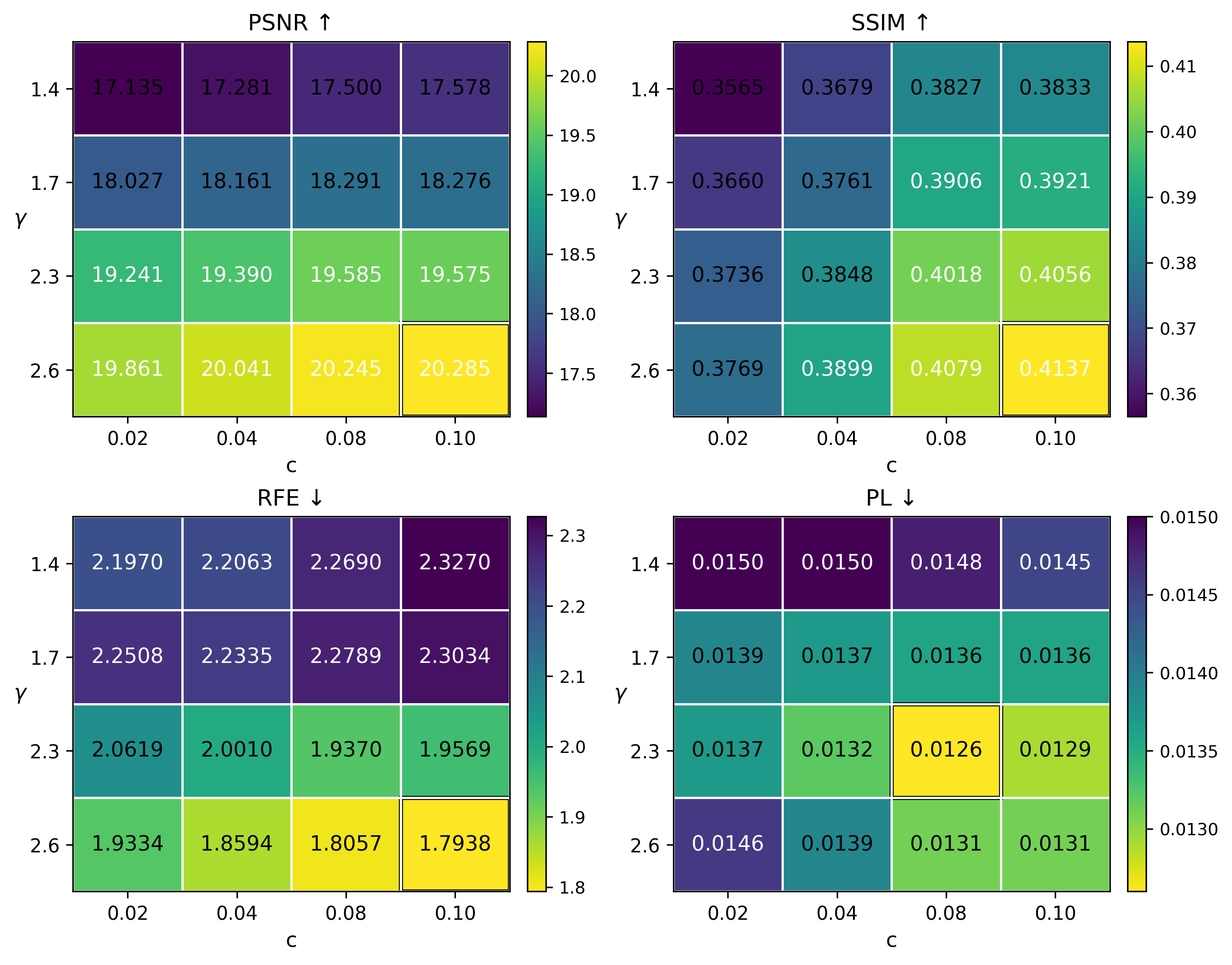}
    }
    \caption{\rev{Sensitivity of reconstruction performance to the simulator parameter combination of $\gamma$ and $C$. Each heatmap shows one evaluation metric over the 2D grid. While different parameter choices in moderate range lead to observable performance changes, the overall variation remains limited, indicating moderate sensitivity and reasonable stability within the tested range.}
    }
\label{fig: gamma_c_heatmap}
\end{figure*}

\begin{figure*}[!t]
    \centering
    \revframe{%
        \includegraphics[width=0.7\textwidth]{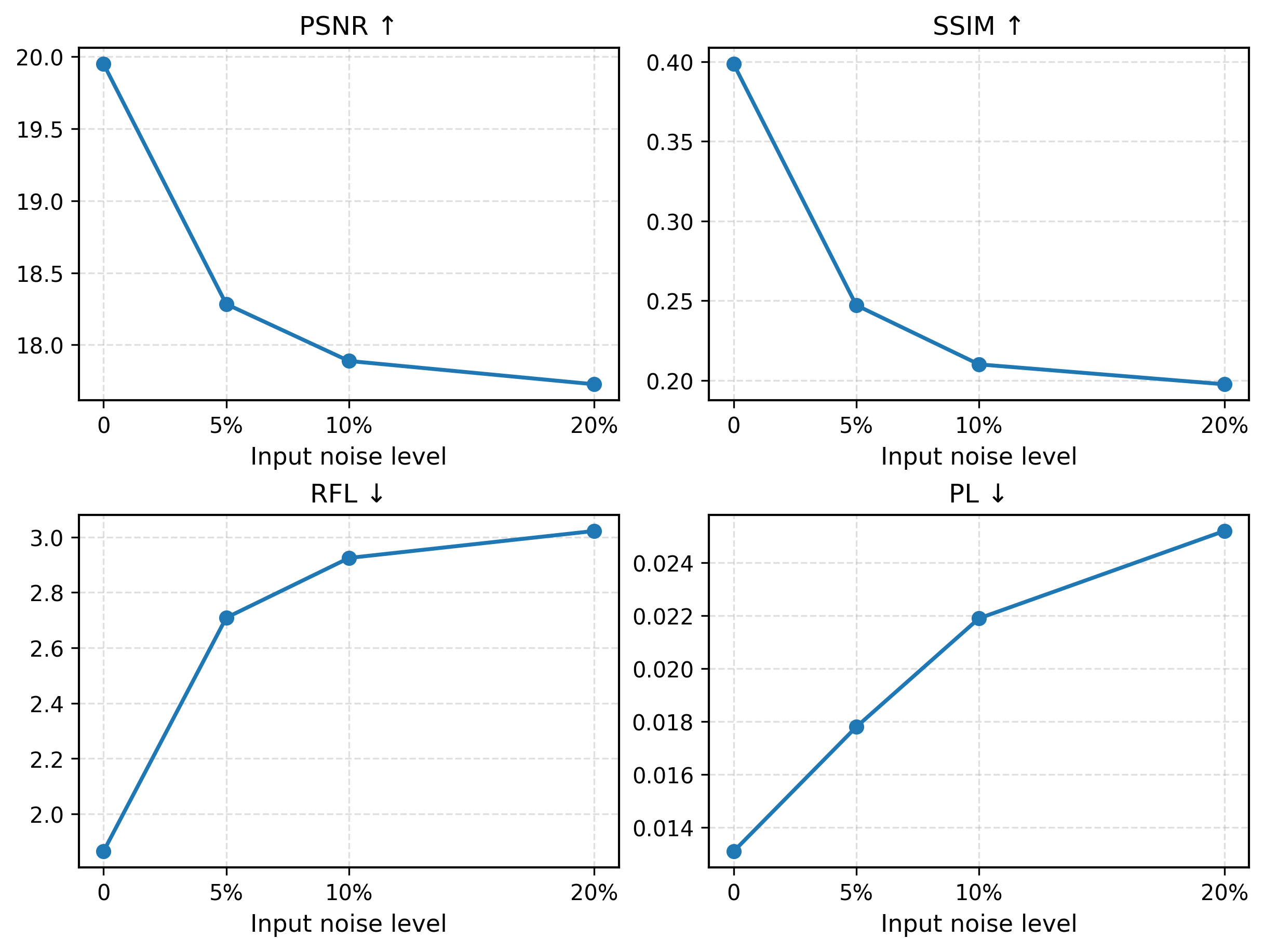}
    }
    \caption{\rev{Degradation in reconstruction quality as the Gaussian noise level increases. The noise percentage denotes the ratio of the noise standard deviation to the maximum image intensity.}
    }
\label{fig: noise_level_impact}
\end{figure*}

\paragraph{Robustness} \rev{To assess the robustness of the pipeline under simulator parameter variations and different noise level, we run a pretrained model on Brain validation set at a $10^\circ$ angular range and a $1^\circ$ angular step with simulator parameters $C \in \left( 0.02, 0.1 \right)$ and $\gamma \in \left(1.4, 2.6 \right)$. Results are summarized in}\revshort{ Fig.~\ref{fig: gamma_c_heatmap}.}\rev{ We also evaluate the performance of the pipeline at different Gaussian noise level, as shown in}\revshort{ Fig.~\ref{fig: noise_level_impact}.} 

\rev{Both parameter variations and increasing noise levels lead to observable changes in PSNR, SSIM, RFE, and PL, indicating that reconstruction quality is influenced by parameter choice and noise. However, the overall variation remains relatively small across the tested range, suggesting that the method is reasonably stable under moderate changes in $\gamma$, $C$ and noise, although it is not entirely insensitive to these factors. Future work will need to focus on refining the methodological design to improve robustness under more challenging and variable imaging conditions.}

%% file: fig/physical_model.tex
\begin{figure*}[!t]
 \centering
   \hspace{-1.2cm}
   \begin{tabular}{ccc}
   \parbox[c]{5cm}{
   \begin{tabular}{ccc}
      \centering
      \includegraphics[width=1.5cm]{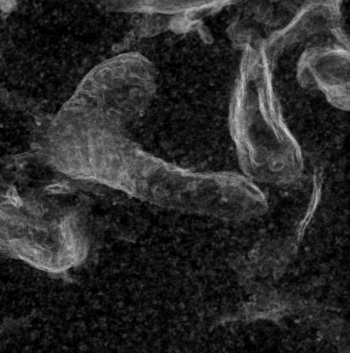}  &
      \includegraphics[width=1.5cm]{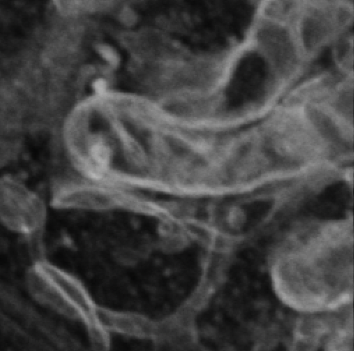}   &
      \includegraphics[width=1.5cm]{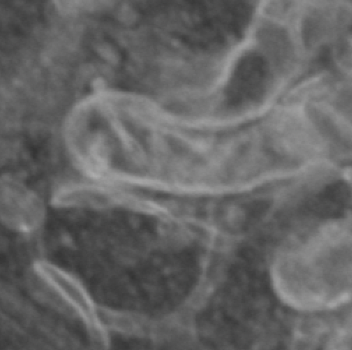}   \\[0.3em] 
      \includegraphics[width=1.5cm]{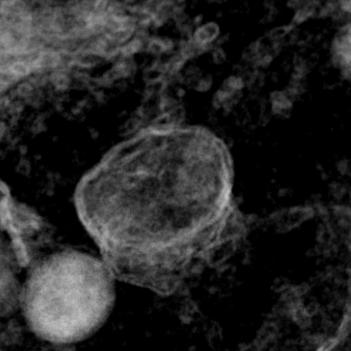}  & 
      \includegraphics[width=1.5cm]{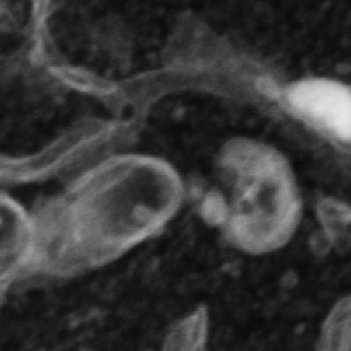}   &
      \includegraphics[width=1.5cm]{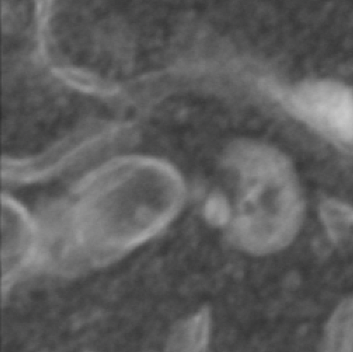}   \\
     \end{tabular}
     }&
    \parbox[c]{5cm}{\includegraphics[height=4cm]{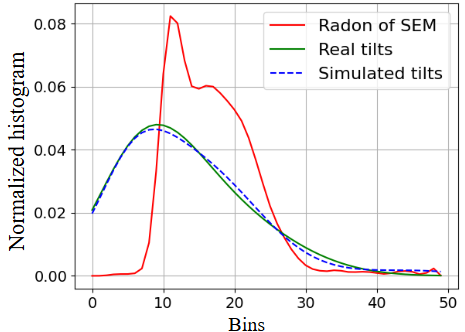}} &
    \parbox[c]{5cm}{
        \revframe{
            \includegraphics[height=4cm]{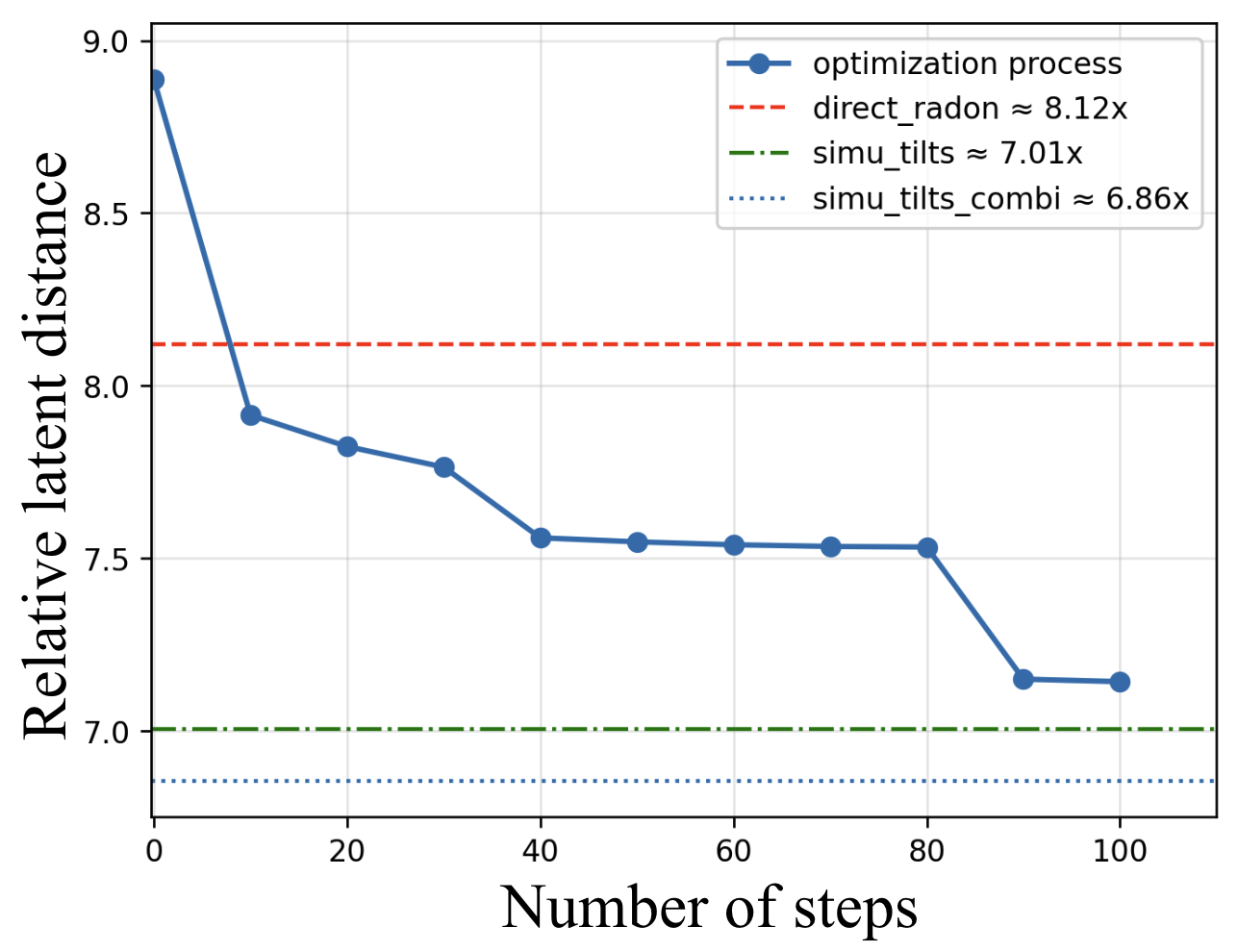}
        }
    } \\
    \hspace{0.8cm}(a) \hspace{1.2cm} (b) \hspace{1.2cm} (c) & (d) & (e)
    \end{tabular}
    \caption{\rev{From FIB-SEM to STEM. (a) Real STEM tilt sequence. (b) Simulated STEM tilt sequence using our proposed simulator. (c) Radon transform of FIB-SEM volume used to generate the synthetic sequence. (d) The corresponding histograms of real STEM, simulated STEM and radon sequence, showing the simulated tilts are very close to the real one. (e) Relative latent distance from real STEM to simulated STEM decreases during parameter adjustment and is smaller than that of radon sequence. If style loss is used in optimization, the relative latent distance can be further decreases to 6.86 (simu\_tilts\_combi)}.}
  \label{fig:sem_to_tem}
\end{figure*}

%% file: fig/results_sem.tex
\begin{table*}[t]
\caption{Performance comparison on four simulated datasets from FIB-SEM.}
\label{tabel: quantitative_results_all}
\centering
\small
\setlength{\tabcolsep}{1pt}
\renewcommand{\arraystretch}{0.92}
\begin{tabular}{llccRR ccRR ccRR}
\toprule
\multirow{2}{*}{Dataset} & \multirow{2}{*}{Algo.} &
\multicolumn{4}{c}{10$^\circ$} & \multicolumn{4}{c}{30$^\circ$} & \multicolumn{4}{c}{60$^\circ$} \\
\cmidrule(lr){3-6} \cmidrule(lr){7-10} \cmidrule(lr){11-14}
& & {PSNR$\uparrow$} & {SSIM$\uparrow$} & {RFE$\downarrow$} & {PL$\downarrow$}
  & {PSNR$\uparrow$} & {SSIM$\uparrow$} & {RFE$\downarrow$} & {PL$\downarrow$}
  & {PSNR$\uparrow$} & {SSIM$\uparrow$} & {RFE$\downarrow$} & {PL$\downarrow$} \\
\midrule
\multirow{7}{*}{Brain}
& SART &   \underline{17.62} & 0.344 & 2.612 & 0.024 &  17.88 & 0.363 & 2.077 & 0.013 & 17.72 & 0.393 & 3.461 & 0.011\\
& \mbir{} & 15.87 & 0.264 & 3.780 & 0.020 &  17.00 & 0.332 & 3.171 & 0.016 & 18.13 & 0.389 & 2.526 & 0.012\\
& \dolc{} & 14.43 & 0.318 & 3.113 & \underline{0.016} & 15.79 & 0.423 & 3.155 & 0.020 & 17.26 & 0.551 & 3.051 & 0.020\\
& \scd{} & 15.19 & 0.296 & 5.000 & 0.024 & 17.36 & 0.427 & 3.235 & 0.015 & 19.38 & 0.524 & 2.031 & 0.007\\
& \ddip{} & 17.57 & 0.332 & \underline{2.550} & 0.020 & \underline{18.98} & 0.437 & \underline{2.028} & \underline{0.011} & 19.68 & 0.571 & 2.641 & 0.010\\
& \pfitre{} & 16.86 & \underline{0.353} & 2.996 & \underline{0.016} & 18.13 & \underline{0.466} & 2.506 & \underline{0.011} & \underline{20.01} & \underline{0.615} & \underline{1.987} & \underline{0.006}\\
& \textbf{\ours{}} & \textbf{19.79} & \textbf{0.490} & \textbf{1.533} & \textbf{0.008} & \textbf{21.07} & \textbf{0.585} & \textbf{1.237} & \textbf{0.006} & \textbf{22.36} & \textbf{0.679} & \textbf{1.000} & \textbf{0.004}\\

\cline{1-14}

\multirow{7}{*}{HeLa 1}
& SART & \textbf{15.62} & 0.293 & \textbf{2.070} & 0.028 & 15.74 & 0.311 & 2.370 & 0.016 & 15.83 & 0.340 & 4.136 & 0.012\\
& \mbir{} & 13.79 & 0.280 & 4.153 & 0.024 & 14.80 & 0.367 & 3.462 & 0.017 & 16.22 & 0.481 & 2.719 & 0.011\\
& \dolc{} & 12.38 & 0.298 & 2.963 & \underline{0.008} & 13.90 & 0.421 & 2.358 & \underline{0.006} & 16.75  & 0.551 & 1.627 & \textbf{0.002}\\
& \scd{} & 14.60 & 0.291 & 3.176 & 0.013 & 15.76 & 0.390 & 2.891 & 0.013 & 17.17 & 0.487 & 2.303 & 0.007\\
& \ddip{} & 15.19 & 0.314 & 2.939 & 0.016 & \textbf{17.58} & 0.451 & 2.400 & 0.011 & \underline{19.09} & 0.561 & 2.052 & 0.006\\
& \pfitre{} & 15.21 & \underline{0.338} & 2.795 & 0.010 & \underline{17.55} & \textbf{0.484} & \underline{1.814} & \textbf{0.004} & \textbf{19.68} & \textbf{0.628} & \textbf{1.373} & \textbf{0.002} \\
& \textbf{\ours{}} & \underline{15.55} & \textbf{0.383} & \underline{2.266} & \textbf{0.006} & 16.70 & \underline{0.477} & \textbf{1.783} & \textbf{0.004} & 17.72 & \underline{0.570} & \underline{1.486} & \underline{0.003} \\

\cline{1-14}

\multirow{7}{*}{HeLa 2}
& SART & \underline{18.73} & \underline{0.387} & 3.149 & 0.019 & \underline{19.04} & 0.410 & \underline{1.919} & 0.010 & 18.97 & 0.444 & 3.710 & 0.009\\
& \mbir{} & 16.81 & 0.334 & 3.338 & 0.021 & 17.77 & 0.419 & 2.847 & 0.015 & 19.25 & 0.529 & 2.276 & 0.010\\
& \dolc{} & 14.28 & 0.343 & 2.939 & \underline{0.014} & 15.20 & \underline{0.433} & 2.458 & 0.012 & 17.62 & 0.568 & \underline{1.853} & 0.006\\
& \scd{} & 17.25 & 0.321 & \underline{2.726} & 0.025 & 18.00 & 0.412 & 2.590 & 0.023 & 19.21 & 0.505 & 2.134 & 0.015\\
& \ddip{} & 17.34 & 0.334 & 2.952 & 0.032 & 18.44 & 0.417 & 2.451 & 0.022 & \underline{19.83} & 0.518 & 1.996 & 0.013\\
& \pfitre{} & 16.61 & 0.326 & 3.160 & 0.019 & 17.77 & 0.430 & 2.535 & \underline{0.009} & 19.42 & \underline{0.573} & 1.956 & \underline{0.005}\\
& \textbf{\ours{}} & \textbf{20.01} & \textbf{0.526} & \textbf{1.476} & \textbf{0.006} & \textbf{21.33} & \textbf{0.615} & \textbf{1.180} & \textbf{0.004} & \textbf{22.46} & \textbf{0.692} & \textbf{0.973} & \textbf{0.003}\\

\cline{1-14}

\multirow{7}{*}{Synapse}
& SART  & 14.36 & 0.393 & 3.323 & 0.040 & 14.33 & 0.412 & 3.347 & 0.025 & 14.22 & 0.452 & 4.051 & 0.019 \\
& \mbir{} & 14.73 & 0.406 & 4.059 & 0.018 & \underline{16.08} & 0.493 & 3.384 & 0.013 & \underline{18.43} & \underline{0.614} & 2.272 & 0.007 \\
& \dolc{} & 11.48 & 0.322 & 4.575 & \underline{0.015} & 13.25 & 0.440 & 3.384 & \underline{0.008} & 15.77 & 0.568 & 3.008 & 0.006\\
& \scd{} & 14.72 & 0.405 & 3.966 & 0.026 & 15.87 & 0.489 & 3.481 & 0.016 & 17.65 & 0.600 & 2.759 & 0.009\\
& \ddip{} & \textbf{15.58} & \underline{0.427} & \underline{3.225} & 0.019 & \textbf{17.11} & \underline{0.514} & \underline{2.540} & 0.012 & \textbf{19.05} & 0.613 & \textbf{1.955} & 0.006\\
& \pfitre{} & 13.42 & 0.321 & 4.719 & 0.017 & 14.84 & 0.413 & 3.750 & 0.009 & 17.40 & 0.553 & 2.507 & \textbf{0.004}\\
& \textbf{\ours{}} & \underline{14.75} & \textbf{0.459} & \textbf{2.984} & \textbf{0.009} & 16.07 & \textbf{0.561} & \textbf{2.382} & \textbf{0.007} & 17.33 & \textbf{0.652} & \underline{1.994} & \underline{0.005}\\
\bottomrule
\end{tabular}
\end{table*}

%% file: fig/results_simtem.tex
\begin{table*}
\caption{Performance comparison on simulated datasets from real data.}
\label{tabel: quantitative_results_real}
\centering
\small
\begin{tabular}{llccRR ccRR}
\toprule
\multirow{2}{*}{Dataset} & \multirow{2}{*}{Algo.} &
\multicolumn{4}{c}{10$^\circ$} & \multicolumn{4}{c}{60$^\circ$} \\
\cmidrule(lr){3-6} \cmidrule(lr){7-10}
& & {PSNR$\uparrow$} & {SSIM$\uparrow$} & {RFE$\downarrow$} & {PL$\downarrow$}
  & {PSNR$\uparrow$} & {SSIM$\uparrow$} & {RFE$\downarrow$} & {PL$\downarrow$} \\
\midrule
\multirow{7}{*}{Mito}
 & SART & 17.20  & 0.337  & 4.190 & 0.020 & 17.12 & 0.348 & 4.566 & 0.011\\
 & \mbir{} & 18.52 & 0.348  & 7.990 & 0.015 & 20.60 & 0.573 & 5.920 & 0.008\\
 & \dolc{} &  17.28 &  0.284 & 7.156 & 0.027 & 17.17 &  0.436 & 6.488 & 0.052\\
 & \scd{} & 18.63 & 0.364 & \underline{2.845} & 0.017 & 20.57 & 0.546 & 2.646 & 0.007\\
 & \ddip{} & \underline{18.74} & \underline{0.375} & 5.556 & 0.019 & \underline{21.42} & \underline{0.616} & \underline{2.355} & \underline{0.005}\\
 & \pfitre{} & 18.21 & 0.328 & 7.540 & \underline{0.013} & 19.36 & 0.506  & 6.934 & 0.010\\
 & \textbf{\ours{}} & \textbf{20.84} & \textbf{0.603} & \textbf{2.279} & \textbf{0.003} & \textbf{24.82} & \textbf{0.854} & \textbf{0.752} & \textbf{0.002}\\
 \cline{1-10}
\multirow{7}{*}{Syna}
 & SART & \textbf{18.36} & \underline{0.331} & \underline{2.074} & 0.025 & 16.95 & 0.372 & 3.989 & 0.012\\
 & \mbir{} & 16.60 &  0.279 & 3.945 & 0.025 & 19.20 & 0.533 & \underline{3.095} & 0.011\\
 & \dolc{} &  14.19 &  0.125 & 5.246 & 0.033 & 16.29 &  0.396 & 4.253 & 0.031\\
 & \scd{} & 16.58 & 0.278 & 5.230 & 0.028 & 18.84 & 0.496 & 3.624 & 0.007\\
 & \ddip{} & 16.87 & 0.301 & 4.492 & \underline{0.020} & \underline{19.42} & \underline{0.536} & 3.102 & \underline{0.005}\\
 & \pfitre{} & 15.40 & 0.195 & 4.008 & 0.023 & 16.08 & 0.296  & 6.604 & 0.023\\
 & \textbf{\ours{}} & \underline{18.17} & \textbf{0.522} & \textbf{1.606} & \textbf{0.007} & \textbf{22.37} & \textbf{0.798} & \textbf{0.708} & \textbf{0.002}\\
\bottomrule
\end{tabular}
\end{table*}

%% file: tex/conc.tex

\section{Conclusion}

We presented \ours{}, a projector-guided conditional 3D diffusion framework for limited-angle electron tomography. The method addresses the missing-wedge challenge by (i) directly operating in 3D volume and train the diffusion model on FIB-SEM data via a physics-motivated simulator (ii) applying projection-consistency projector at every reverse-diffusion step. Across mitochondria and synapse datasets, \ours{} consistently improved reconstruction quality over iterative method SART, U-Net based artifact suppression baseline \pfitre{}, and recent diffusion-based baselines \mbir{}, \dolc{}, \ddip{} and \scd{} under severe angular restrictions. The gains were most pronounced at $10^\circ$ angular range, where slice-wise approaches struggle to maintain volumetric consistency or have degraded quality. On real-data-derived tests, \ours{} balanced fidelity and artifact suppression better than competing methods. Finally, on raw real tilt series collected with angular range as narrow as $8^\circ$ ($1^\circ$ or $2^\circ$ steps), \ours{} produced clear and plausible reconstructions with enhanced contrast and structural coherence, demonstrating practical utility for tilting-range-constrained acquisitions.

\subsection{Limitations} \ours{} relies on an approximate simulator to bridge FIB-SEM and TEM, and residual domain gaps due to specimen-specific physics and resolution differences remain. In addition, mechanical inaccuracies in the electron microscope can further limit performance and degrade reconstruction quality, even when uncertainty weighting is employed.

\subsection{Future work} We plan to (i) further narrow the simulator gap via \rev{testing-time adaptation strategy and 3D-aware domain adaptation, (ii) jointly estimate alignment, simulator parameters and reconstruction within the diffusion-projector loop.}
\newline\newline
In summary, \ours{} provides a practical path to high-quality 3D reconstructions from very limited tilt ranges by transferring structural priors from accessible volumetric data and tightly coupling them with consistency projector. We believe this capability can broaden the use of limited-angle rapid electron tomography in biological and materials studies.